\documentclass[10pt]{article} 
\usepackage{arxiv_preprint}

\usepackage{amsmath,amsfonts,bm}

\def\eqref#1{equation~\ref{#1}}

\def\1{\bm{1}}

\DeclareMathAlphabet{\mathsfit}{\encodingdefault}{\sfdefault}{m}{sl}
\SetMathAlphabet{\mathsfit}{bold}{\encodingdefault}{\sfdefault}{bx}{n}

\def\gA{{\mathcal{A}}}

\def\gE{{\mathcal{E}}}

\def\gG{{\mathcal{G}}}

\def\sC{{\mathbb{C}}}

\def\sM{{\mathbb{M}}}

\def\sO{{\mathbb{O}}}

\usepackage{hyperref}
\usepackage{url}
\usepackage{enumitem}
\usepackage{booktabs}
\usepackage{tabularx}  
\usepackage{ragged2e} 
\usepackage{array}     
\usepackage{xcolor}    
\usepackage{multirow,colortbl}
\usepackage{graphicx}
\usepackage{caption}
\usepackage{subcaption}
\usepackage{cleveref}
\usepackage{wrapfig}
\usepackage{mathtools}

\title{Drawing with Strangers: Population Scaling Drives Zero-Shot Mutual Intelligibility in Emergent Sketching}

\author{\name Jooyeon Kim \email jooyeon.kim@unist.ac.kr \\
      \addr Graduate School of Artificial Intelligence\\
      UNIST
       }

\begin{document}

\maketitle

\begin{abstract}
Generalization in emergent communication has largely focused on novel inputs or linguistic structures, yet the capacity for agents to communicate with strangers from strictly disjoint communities remains relatively unexplored.
In this work, we formalize this capability as \textit{zero-shot mutual intelligibility (ZMI)}: successful communication between independently trained populations without prior exposure. 
Leveraging emergent sketching---in which agents communicate through sets of drawn strokes---as a visually grounded modality, we find that scaling the training population substantially improves ZMI across independent groups.
Crucially, as we scale the population size, in-group communicative variation increases, preventing co-adaptation into homogeneity.
Simultaneously, cross-group variation decreases, indicating a structural convergence toward a certain type of universality.
Further analysis reveals that this universality is achieved through perceptual grounding: scaled populations increasingly anchor their emergent sketches on the objective visual resemblance of the target images.
Together, these results position ZMI as a distinct axis of generalization in emergent communication and suggest a route toward socially interoperable artificial agents.
\end{abstract}

\section{Introduction}

A defining characteristic of social intelligence is the capacity to coordinate with others outside of one's immediate cohort.
In biological societies, the evolutionary success of humans is often attributed to this ability—specifically, the establishment of robust communication protocols that translate seamlessly to individuals or groups outside of immediate kin \citep{nowak1999evolution, tomasello2010origins}.
This universality, the ability to exchange information with complete strangers without a pre-shared dictionary, is a cornerstone of large-scale cooperation and social organization \citep{skyrms2010signals}.

In artificial intelligence, the field of emergent communication investigates how autonomous agents can develop communication protocols from scratch, driven purely by task-level coordination pressure~\citep{foerster2016learning,sukhbaatar2016learning,lazaridou2017multi}.
This paradigm implies that pragmatic language arises not from pre-defined rules, but as a functional solution to coordination problems.
While significant progress has been made in analyzing these protocols (refer to \citet{boldt2024a} for the review), evaluations of their generalizability have largely centered on the agents’ descriptive capabilities.
Concretely, prior work characterizes generalizability along two primary dimensions: input-level generalization, testing agents’ ability to describe novel stimuli both in- and out-of-distribution~\citep{mu2021emergent,rita2022emergent}, and language-level generalization, evaluating if emergent protocols exhibit human-like structural properties such as systematicity~\citep{hill2020environmental} and compositionality~\citep{chaabouni2020compositionality,carmeli2024concept}.

In this work, we identify and formalize a distinct regime of generalization: \emph{zero-shot mutual intelligibility (ZMI)}, defined as the ability to successfully communicate with agents initialized and trained in strictly disconnected communities. This formulation draws direct inspiration from the problem of zero-shot coordination (ZSC) in multi-agent cooperation \citep{hu2020other}. However, while ZSC primarily addresses the alignment of joint policies, ZMI focuses specifically on the semantic alignment of the communication protocol itself. The core challenge is ensuring that a signal generated by one group carries a preserved meaning to strangers without prior exposure. This form of social generalization remains largely underexplored in the emergent communication literature, as existing symbolic approaches often devolve into idioglossia — co-adapted protocols that are unintelligible to outsiders \citep{lu2020countering,michelrevisiting}.

We proceed to demonstrate, to the best of our knowledge, the first successful realization of high-fidelity ZMI between independent, disjoint AI populations. To achieve this, we leverage the modality of emergent sketching \citep{mihai2021learning}. Unlike symbolic language protocols that often rely on arbitrary signs of token mappings and rules, sketches communicate iconically through visual resemblance and spatial structure, thereby anchoring meaning in perceptual reality~\citep{fay2014iconicity,garrod2007foundations}.
We provide empirical evidence that this visually grounded communication channel enables independently trained agents sampled from disjoint populations to achieve high mutual intelligibility and communication success.

Crucially, we identify population scale, combined with a visually grounded modality, as an effective driver for this emergence. While prior studies in symbolic emergent communication suggest that increasing group size yields diminishing returns  \citep{chaabouni2022emergent}, parallels can be drawn from the multi-agent cooperation literature where population-based training promotes robustness~\citep{jaderberg2017population,strouse2021collaborating}.
We observe convincing evidence that scaling the communication cohort consistently improves ZMI. In fully connected settings, achieving a baseline communication accuracy requires total training iterations that scale linearly with population size, despite quadratically increasing connections. This efficiency indicates that agents avoid the superlinear cost of memorizing partner-specific quirks. These results lend support to scaling social interaction as a mechanism for promoting universally intelligible conventions.

To investigate the mechanism by which population scaling improves ZMI, we first analyze how communicative variance shifts with group size.
As populations scale, in-group variation increases gradually, acting as a regularizer that prevents co-adaptation into isolated, homogeneous idioglossia.
Consequently, the inability to rely on partner-specific quirks forces populations toward a certain type of communicative universality, driving a substantial decrease in cross-group variation.
At the largest scales, the gap between in-group and cross-group variation diminishes significantly, indicating convergence toward shared protocols rather than arbitrary local dialects.
Furthermore, we find this universality is inherently tied to perceptual grounding: higher ZMI strongly correlates with greater visual resemblance between emergent sketches and target images.
Thus, population scaling improves cross-group intelligibility by anchoring communication on the intrinsic visual features of the target images, naturally leading to generalization across strangers.

Our contributions are threefold:
\begin{enumerate}[leftmargin=*]
    \item We formally introduce zero-shot mutual intelligibility (ZMI), the ability of independently trained agent populations to successfully communicate without prior exposure. This formulation extends existing notions of generalization in emergent communication beyond descriptive or structural metrics, establishing social interoperability across populations as a distinct and measurable objective.
    \item We provide empirical evidence that high-fidelity ZMI emerges between disjoint populations through emergent sketching. Specifically, fully connected population scaling improves ZMI at approximately linear training cost, proving that maintaining mutual compatibility across diverse partners effectively suppresses partner-specific co-adaptation that intrinsically require superlinear training iterations.
    \item We investigate the underlying mechanisms behind this scaling-induced improvement in ZMI by analyzing shifts in communicative variance and perceptual grounding. We show that scaling increases in-group diversity while decreasing cross-group variation, compelling independently trained groups to abandon arbitrary drift and converge on universally shared, perceptually grounded conventions.
\end{enumerate}

Our work sits at the intersection of three lines of research: scaling laws in emergent communication, which study how population size and diversity shape protocol formation \citep{graesser2019emergent,kim2021emergent, chaabouni2022emergent, michelrevisiting}; population-based training in multi-agent reinforcement learning, which shows that diverse social exposure promotes robustness and generalization \citep{jaderberg2017population, strouse2021collaborating}; and visually grounded sketch-based communication, which provides an iconic, perceptually anchored signaling modality~\citep{mihai2021learning}.
By bringing together these perspectives, we take a step toward emergent communication systems that extend beyond closed-population protocol optimization toward broader social compatibility, moving toward agents that can communicate with unseen partners. 
\clearpage
\section{Methodology}
This section describes the methodological framework used to study the emergence of sketch-based communication and cross-group generalization. We first introduce the referential game and its differentiable sketch-based variant, and then present our population-based training formulation together with the zero-shot mutual intelligibility (ZMI) metric used to evaluate generalizability in cross-group communication settings.
\subsection{Background}
\paragraph{Referential game.}
We consider a referential game~\citep{lazaridou2017multi,havrylov2017emergence}, a modern instantiation of the Lewis signaling game \citep{lewis1969convention}, commonly used to study the emergence of communication protocols between interacting agents.
Let $\sO$ denote a space of objects and $\sM$ a space of messages.
In each episode, a target object $t \in \sO$ is sampled from $\sO$ and revealed only to the sender.
Based on the target object, the sender produces a message $m \in \sM$ intended to signify the target.
The receiver is then presented with a candidate set
$\sC = \{o_1, \dots, o_K\} \subset \sO$,
containing exactly one target object and $K-1$ distractor objects, randomly permuted.
After observing both the message and the set of candidates, the receiver selects an index
$\hat{i} \in \{1,\dots,K\}$,
corresponding to its prediction of the target.
Communication is successful if the selected candidate corresponds to the target object, i.e.,
$o_{\hat{i}} = t$.
A communication protocol is established by a sender mapping from objects to messages and a receiver mapping from messages and candidate sets to candidate indices over repeated episodes during the training process.

\paragraph{Sketch-based referential game.}
We extend the referential game framework to a sketch-based communication setting, inspired by~\citet{mihai2021learning}. In this variant, both sender and receiver agents are parameterized by deep neural networks, and communication occurs through sketches rather than symbolic messages.
Let $\sO$ denote a set of RGB images representing objects. In each episode, a target image $t \in \sO$ is presented to the sender. The sender processes $t$ and outputs a sequence of $D$ stroke parameters, each defined by a 4-dimensional vector specifying the start and end coordinates $(x_{\text{start}}, y_{\text{start}}, x_{\text{end}}, y_{\text{end}})$ of a line segment. These vectorized strokes are then rasterized into an RGB image $m \in \mathcal{M}$ using a differentiable rendering process based on Gaussian products~\citep{mihai2021differentiable}, allowing gradients to flow through the drawing procedure during training (refer to \autoref{app:rasterization} for details).
The receiver is presented with the sketch $m$ and a candidate set $\sC = \{o_1, \dots, o_K\} \subset \sO$, which includes the original target image and $K-1$ distractor images, randomly ordered. The receiver processes $m$ and $C$ to produce a probability distribution over the candidate indices and selects an index $\hat{i} \in \{1, \dots, K\}$.
Communication is deemed successful if $o_{\hat{i}} = t$.

\subsection{Emergent sketching across scales and zero-shot mutual intelligibility}

We study the emergence of sketch-based communication protocols in a population setting consisting of multiple independently interacting communication groups.
Let there be $M$ communication groups, each containing $N$ agents.
For clarity of exposition, we assume a fixed group size and structure across all groups, though our formulation naturally extends to heterogeneous group sizes and connectivities.
Each agent is equipped with both a sender module and a receiver module and may assume either role in a given episode.
During training, communication occurs only between agents belonging to the same group; no cross-group interaction is permitted.
Note that our formulation reduces to an undirected, bipartite sender--receiver graph with $N$ sender modules and $N$ receiver modules when self-communication is allowed.

\paragraph{Intra-group communication structure.}
Within each group, communication interactions are governed by a communication graph.
For group $g \in \{1,\dots,M\}$, we define a graph
$\gG_g = (\gA_g, \gE_g)$,
where $\gA_g$ denotes the set of agents in group $g$ and $\gE_g \subseteq \gA_g \times \gA_g$ denotes allowed sender--receiver communication pairs.
We assume that each communication graph is connected, i.e., there are no disjoint subgroups and every pair within a group has a path.
Different connectivity structures are possible, ranging from sparse structured topologies to fully connected communication, where any agent may directly communicate with all others.
For the ease of exposition, we limit the scope to a fully connected setting in this research.

\paragraph{Zero-shot mutual intelligibility.}
We evaluate cross-group communication using zero-shot mutual intelligibility (ZMI), which measures the ability of independently trained agents to successfully communicate without prior exposure to each other.
Let $a_n^{(g)}$ denote the $n$-th agent in group $g$.
For two agents belonging to different groups, $a_n^{(g)}$ and $a_{n'}^{(g')}$ with $g \neq g'$, we define individual zero-shot mutual intelligibility as the communication success rate when $a_n^{(g)}$ acts as the sender and $a_{n'}^{(g')}$ acts as the receiver.
ZMI is directional: communication from group $g$ to $g'$ is evaluated separately from communication in the reverse direction.
Note that the term mutual intelligibility is originally suggested in the socio-linguistics literature and is also asymmetric~\citep{gooskens2007contribution,gooskens2018mutual}.

Specifically, ZMI between a pair of agents is defined as
\begin{equation}
\text{ZMI}_{\text{pair}}(a_n^{(g)} \rightarrow a_{n'}^{(g')})
\coloneq
P\big(o_{\hat{i}} = t \;\big|\; 
\text{sender} = a_n^{(g)}, \;
\text{receiver} = a_{n'}^{(g')} ; \; g \neq g' \big ),
\label{pair-zmi}
\end{equation}
where communication success is measured under the sketch-based referential game described above.

Individual-level ZMI is computed by averaging out all agents outside of $g$:
\begin{equation}
\text{ZMI}_{\text{indiv}}(a_n^{(g)})
=
\frac{1}{(M-1)N}
\sum_{\substack{g' = 1 \\g' \neq g}}^{M}
\sum_{n'=1}^N
\text{ZMI}_{\text{pair}}(a_n^{(g)} \rightarrow a_{n'}^{(g')}),
\label{indiv-zmi}
\end{equation}

and ZMI between a pair of groups is computed as:
\begin{equation}
\text{ZMI}_{\text{group}}(g \rightarrow g')
=
\frac{1}{N^2}
\sum_{n = 1}^{N}
\sum_{n'=1}^N
\text{ZMI}_{\text{pair}}(a_n^{(g)} \rightarrow a_{n'}^{(g')}).
\label{group-zmi}
\end{equation}

Finally, unless otherwise specified, ZMI refers to the population-level measure obtained by averaging individual ZMI across all cross-group sender--receiver pairs:
\begin{equation}
\text{ZMI}
=
\text{ZMI}_{\text{pop}}
=
\frac{1}{M(M-1)N^2}
\sum_{g \neq g'}
\sum_{n=1}^N
\sum_{n'=1}^N
\text{ZMI}_{\text{pair}}(a_n^{(g)} \rightarrow a_{n'}^{(g')}).
\label{population-zmi}
\end{equation}
This metric captures the extent to which independently trained communication groups, under identical size and connectivity, converge toward mutually intelligible sketching conventions.

\section{Experimental setup}
\paragraph{Model configurations.}
Throughout the paper, we use convolutional neural network (CNN) architectures~\citep{lecun1998gradient,krizhevsky2012imagenet} to process pixel-based RGB images.
Refer to \autoref{tab:model_architecture}, \autoref{app:training_details}, and \autoref{app:compute_infrastructure} for the model configurations, training details and computational infrastructure.

\paragraph{Sender.}
For the sender module, a target image $t$ is processed by a CNN module, producing a flattened feature representation.
The feature vector is passed through a fully connected layer followed by a sigmoid activation.
The resulting output is reshaped into a $2 \times 2 \times D$ tensor, where $D$ denotes the number of strokes generated per target image.
Each stroke is parameterized by its start and end coordinates in normalized image space.
The resulting set of stroke parameters is rasterized using the differentiable Gaussian product renderer~\citep{mihai2021differentiable}, producing an RGB sketch image that serves as the communication message.

\paragraph{Receiver.}
The receiver module processes both the rasterized sketch message and the candidate images to produce a score over candidate indices.
The rasterized sketch image is first processed using a CNN-based encoder, producing a fixed-dimensional embedding vector.
Similarly, each candidate image is processed independently through a separate CNN-based encoder to produce candidate embeddings in the same feature space.
For a candidate set of size $K$, the sketch embedding is broadcast across candidates and concatenated with each candidate embedding to form joint sketch--candidate representations.
These joint representations are passed through a multi-layer perceptron (MLP), producing a scalar logit for each candidate.
The resulting logits are interpreted as scores over candidate indices, from which the predicted target index $\hat{i}$ is obtained.
This architecture allows the receiver to learn joint compatibility between sketch messages and candidate images through learned nonlinear interactions.\footnote{
Alternatively, the final logits may be computed via an inner product between sketch and candidate embeddings. In practice, we observed no significant differences between these design choices and therefore adopted the MLP-based scorer by default.}

\paragraph{Training configuration.}
All models are trained using the Adam optimizer~\citep{kingma2014adam} with a learning rate of $10^{-4}$ and a batch size of 128.
Target and candidate images are normalized to have mean and standard deviation of 0.5.
Each agent maintains independent sender and receiver parameters with separate optimizers, resulting in $2N$ optimizers per communication group.
During training, each iteration uniformly samples $B$ communication edges from the intra-group communication graph $\mathcal{E}_g$.
For each sampled edge, a target image and a set of candidate images are drawn.
Unless otherwise specified, the number of candidates is fixed to $K = 10$.
Model parameters are optimized using the cross-entropy loss computed over candidate logits.
Training proceeds until the early stopping criterion described below is met.

\paragraph{Training termination and fair ZMI evaluation.}
\label{para_training_termination}
A principled training termination criterion is essential to ensure that ZMI reflects the intrinsic generalization properties induced by communication structure, rather than artifacts of optimization budget or training exposure.
In particular, we aim to ensure that the ZMI values are not affected by insufficient or excessive training.

A naive approach is to fix the number of gradient updates per sender and receiver module across all group sizes.
However, as group size increases, each sender must successfully communicate with an increasing number of receivers (and vice versa).
Under a fixed update budget, this reduces effective communication exposure per sender--receiver pair, leading to systematic underfitting in larger groups.
For example, in fully connected settings, this corresponds to total training iterations scaling approximately linearly with group size, while the number of interaction partners per module also increases linearly.

Conversely, one can fix training exposure at the level of communication edges by scaling the total number of training iterations proportionally to the number of edges.
In fully connected settings, the number of communication edges scales quadratically with group size.
In this case, each sender and receiver module undergoes a rapidly increasing number of gradient updates as group size grows.
This leads to excessive optimization of individual modules relative to smaller-group settings, increasing the tendency to overfit to training interactions and for the larger groups.

To balance these effects, we adopt an early stopping criterion based on reference validation performance.
We construct a held-out validation set and measure communication accuracy under in-group communication, i.e., communication with connected agents.
Training is terminated once this reference validation accuracy is reached, with the evaluation window scaled linearly with group size to account for slower convergence in larger populations.
In our work, the communication success metric using a validation set is used as a reference signal to align training across configurations, whereas prior work, such as \citet{chaabouni2022emergent}, uses this held-out in-distribution communication accuracy as a generalization metric itself.\vspace{-1mm}

\paragraph{Target variation and semantic generalization.}
An important design choice in referential communication games concerns how the target image is presented to the sender and receiver. The most basic setting uses identical target images for both agents, allowing communication to be grounded in instance-level visual correspondence. In this work, we instead adopt a target-variation setting, in which the sender and receiver observe different images that belong to the same semantic class, e.g., different instances of the same object category or concept. This setting encourages agents to communicate at the level of shared semantic attributes rather than pixel-level identity. Prior work has shown that such variation is critical for inducing compositional and semantically grounded communication protocols in emergent language settings \citep{choi2018compositional,lazaridou2020emergent,mu2021emergent}, and in emergent sketch-based communications \citep{mihai2021learning}. Establishing a formal connection between semantic generalization that arises from perceptual variation and social generalization that arises from cross-population interoperability, as captured by ZMI, represents a meaningful direction for future work.
\clearpage

\section{Results}
In this section, we analyze the relationship between population scaling and ZMI among artificial agents communicating through emergent sketching. 
In \autoref{sec:scale}, we demonstrate that scaling a fully connected communication graph substantially improves ZMI while maintaining an approximately linear training cost. 
We show that this population scaling leverages the mutual pressure of decentralized, bidirectional in-group communicative diversity, which prevents co-adapted homogeneity and encourages disjoint groups to converge toward a certain type of universality.
In \autoref{sec:perceptual}, we reveal that such universality via population scaling is achieved by increasingly anchoring their emergent sketches in the perceptual features of the target images. 
Finally, we note that the ZMI improvements driven by population scaling represent a robust phenomenon consistent across various experimental settings (e.g., candidate sizes, stroke counts, reference in-group validation communication accuracies) and datasets (\autoref{app:celebA}), and confirm that simply increasing individual model capacity yields no comparable improvement. 
Due to the space limit, the comprehensive robustness analyses and model size ablations are detailed in \autoref{app:robustness} and \autoref{app:model_scaling_ablation}, respectively. \vspace{-1mm}

\subsection{Population scaling and zero-shot mutual intelligibility}
\begin{figure}[t]
    \centering
    \begin{subfigure}[b]{\textwidth}
        \centering
        \includegraphics[width=0.8\textwidth]{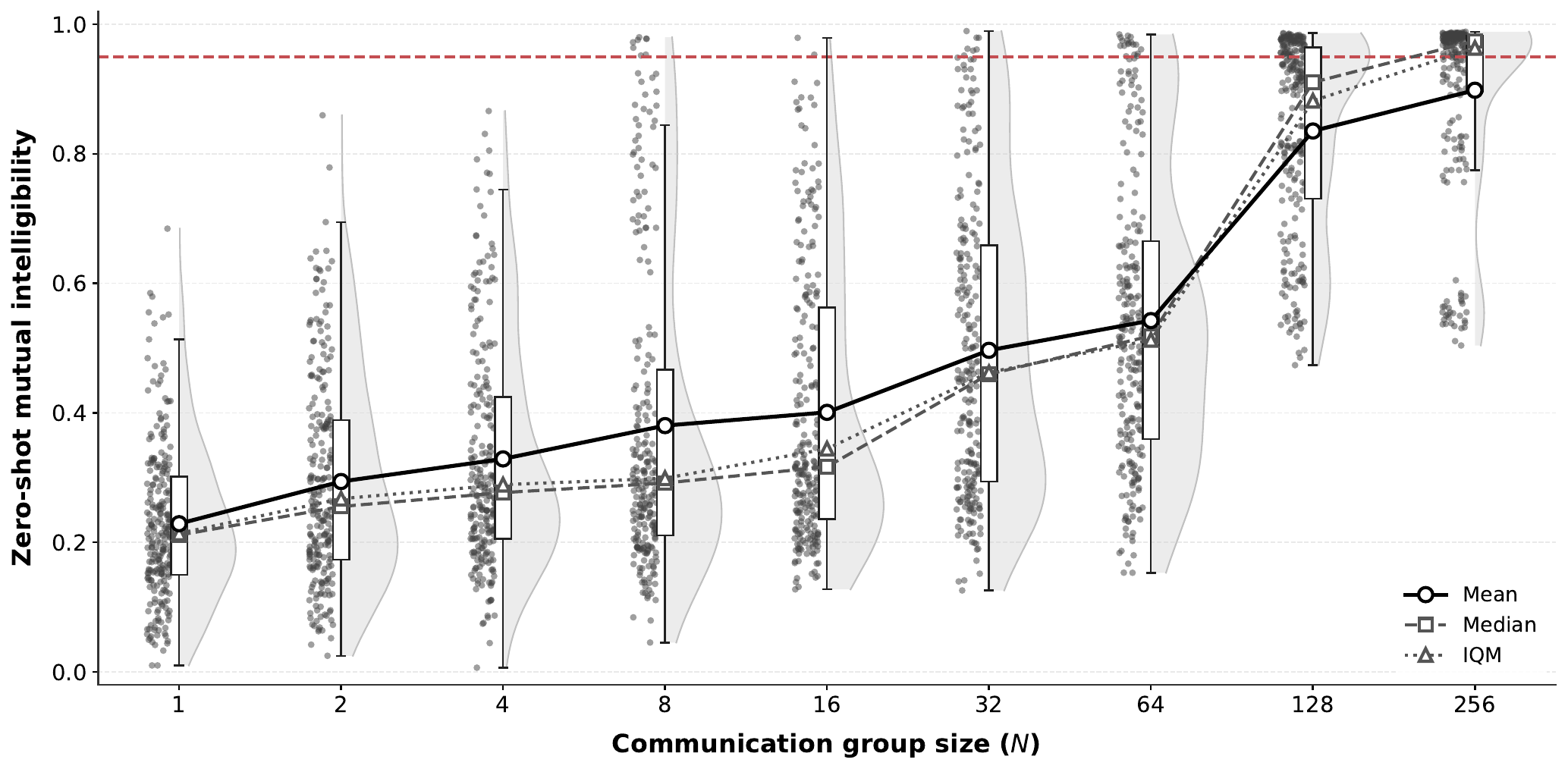}        
        \caption{MNIST dataset}
    \end{subfigure}
    \hfill
    \begin{subfigure}[b]{\textwidth}
        \centering
        \includegraphics[width=0.8\textwidth]{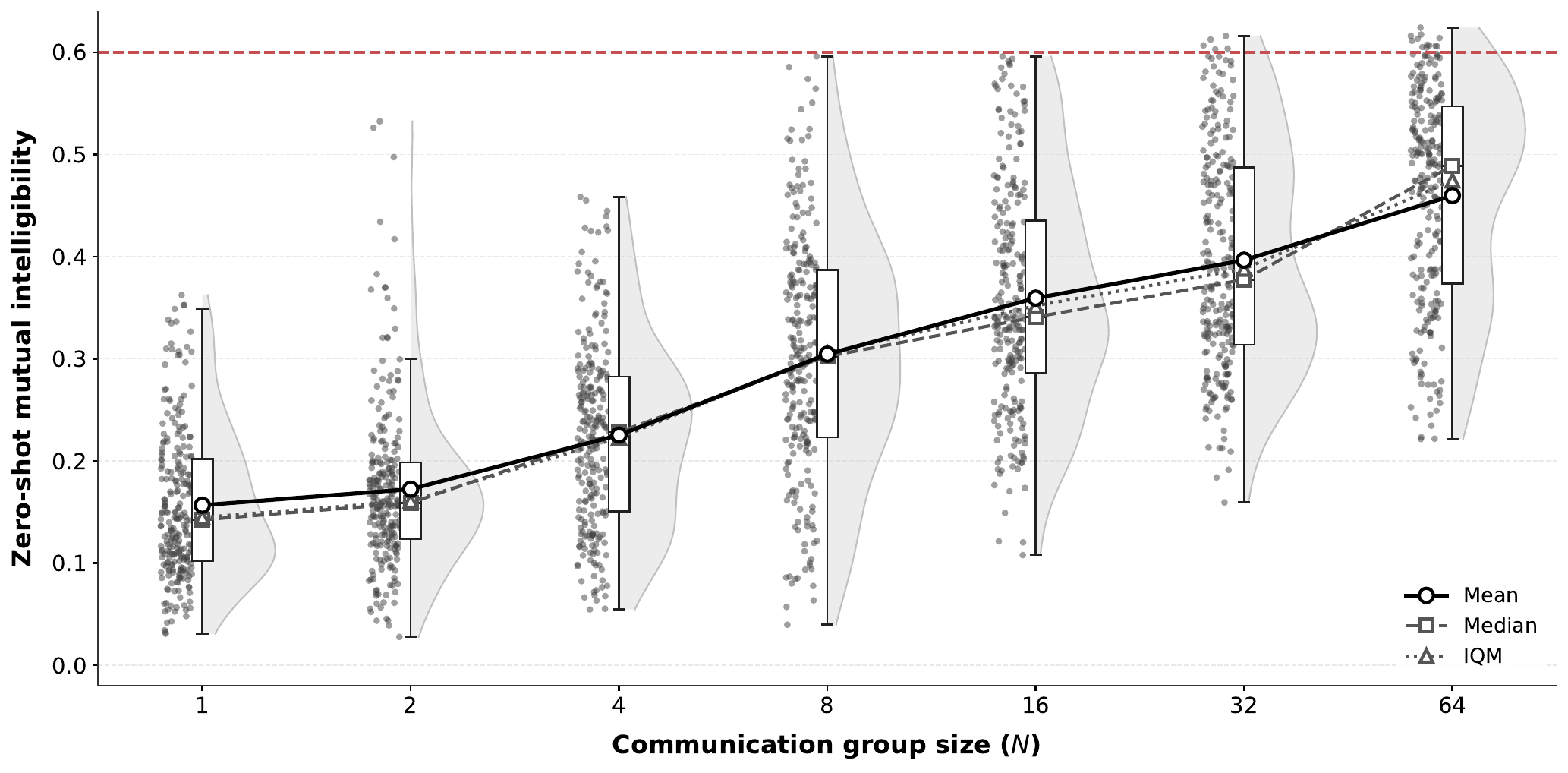}        
        \caption{CIFAR-10 dataset}
    \end{subfigure}
\caption{
    Zero-shot mutual intelligibility (ZMI) evaluated on two datasets.
    Each dot represents group-level ZMI (\autoref{group-zmi}).
    The number of groups $M$ is set to 16, thereby the number of dots is 240 ($16 \times 15$) for each population scale.
    The solid line represents the population-level ZMI (\autoref{population-zmi}), along with median (dashed) and interquartile mean (IQM; dotted).
    For both datasets, ZMI increases substantially by scaling the communication group size ($N$), gradually getting closer to the in-group communication accuracy, which is represented as red, dashed horizontal line, set to 0.95 for MNIST and 0.6 for CIFAR-10.
    }
\label{fig:zmi}
\end{figure}

\vspace{-1mm}
\label{sec:scale}
In \autoref{fig:zmi}, we observe a relationship between communication group size ($N$) and ZMI, evaluated on the MNIST~\citep{lecun1998gradient} and CIFAR-10~\citep{krizhevsky2009cifar} datasets.
For both datasets, we use the default dataset configurations, including dataset size, image resolution, and the training/validation splits.
For both datasets, we evaluated ZMI across population sizes that increase as powers of 2.
Due to the time constraint, we evaluated up to the population size of 256 for the MNIST dataset and 64 for the CIFAR-10 dataset.
ZMI is measured after training is terminated, i.e., once the in-group validation communication accuracy reaches 0.95 for MNIST and 0.60 for CIFAR-10, respectively.
Each dot represents group-level ZMI: $\text{ZMI}_{\text{group}}$.
As $N$ increases, it is computationally prohibitive to compute \autoref{group-zmi} for all possible pairs from disjoint groups $g$ and $g'$.
Therefore, we employ a sampling approach in which we repeatedly estimate $\text{ZMI}_{\text{pair}}$ (\autoref{pair-zmi}) using target image samples of size $256$ for randomly selected sender–receiver pairs $(n, n')$, drawn from groups $g$ and $g'$, respectively.
This process continues until the sample mean converges within a $1\%$ error margin at a $99\%$ confidence level.
Throughout, we assess ZMI with respect to 16 disjoint groups, resulting in a total of 240 (16 $\times$ 15) dots, each of which represents $\text{ZMI}_{\text{group}}(g \rightarrow g')$.\vspace{-1mm}

We observe that scaling the communication group size consistently improves ZMI, in terms of the mean, median, and interquartile mean (IQM).
At $N = 1$, ZMI starts at $0.229$ and $0.157$ for MNIST and CIFAR-10, respectively.
Note that for both datasets, the baseline communication accuracy is $0.1$, as the number of candidates $K$ is set to $10$; this result suggests the sketches created in the single-agent, self-communication setting still generalize better than the random baseline. 
ZMI increases substantially at the largest population size.
For MNIST, at $N=256$, the mean, median, and IQM increase up to 0.898, 0.974, and 0.964, respectively, given the in-group communication accuracy of 0.95 achieved during the training process.
For CIFAR-10, at $N=64$, the mean, median, and IQM increase up to 0.460, 0.489, and 0.474, respectively, given the in-group communication accuracy of 0.60 achieved during the training process.\vspace{-1mm}

\begin{figure}[t]
    \centering
    \begin{subfigure}[b]{0.3\textwidth}
        \centering
        \includegraphics[width=\textwidth]{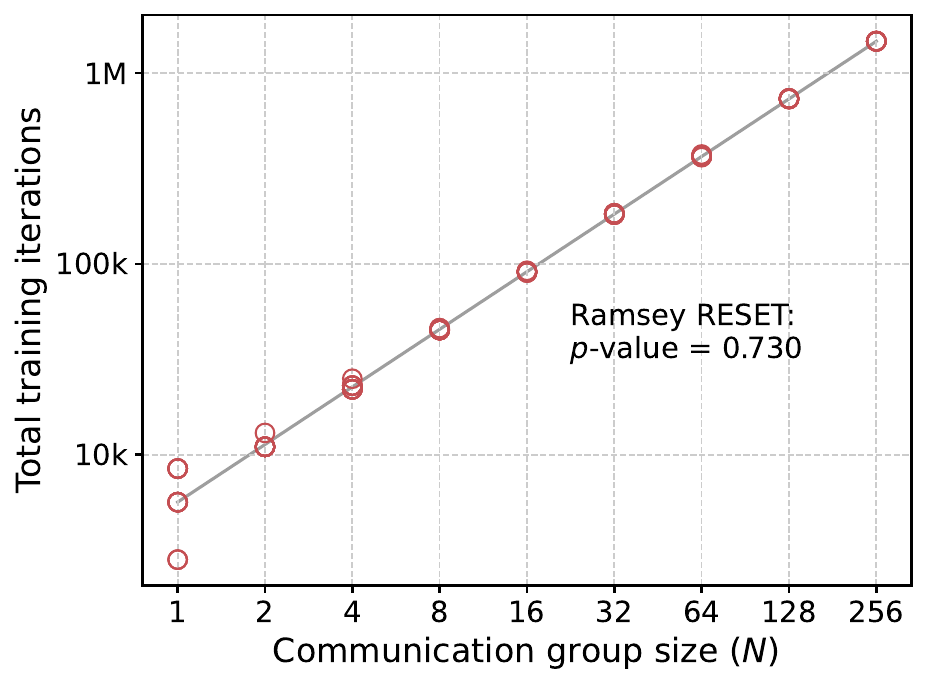}        
        \caption{MNIST dataset}
    \end{subfigure}
    \hspace{0.05\textwidth}
    \begin{subfigure}[b]{0.3\textwidth}
        \centering
        \includegraphics[width=\textwidth]{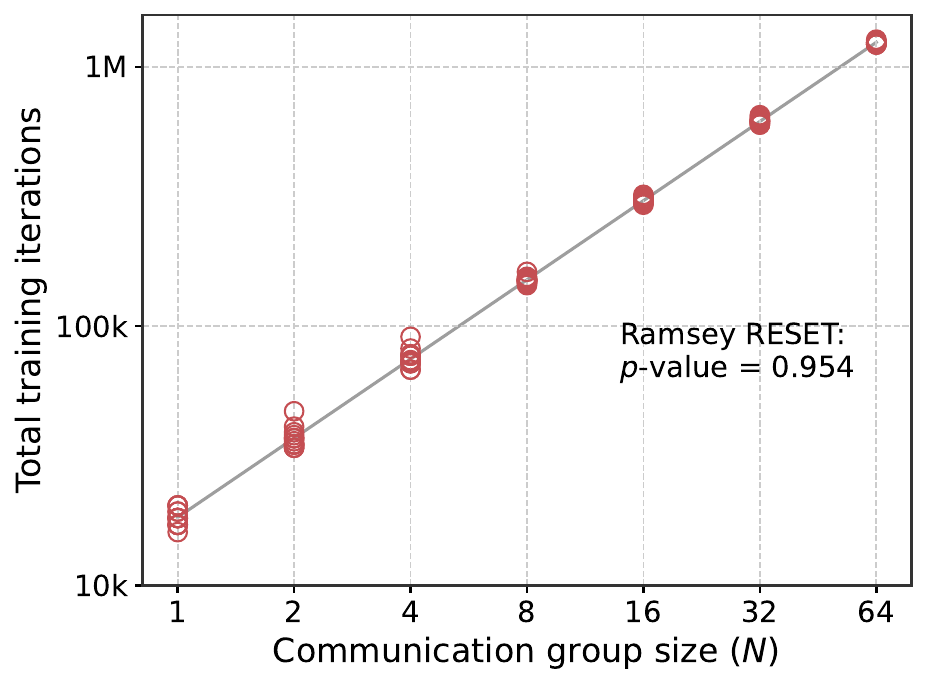}        
        \caption{CIFAR-10 dataset}
    \end{subfigure}
\caption{
    Population scaling increases training cost approximately linearly.
    For each communication group size $N$, we train $M=16$ independently initialized groups until they reach a fixed benchmark in-group validation accuracy.
    Despite the quadratically increasing number of possible sender--receiver pairings in a fully connected population, the total number of training iterations grows approximately linearly with $N$ for both MNIST and CIFAR-10.
    %
    %
    This subquadratic scaling suggests that larger populations do not rely on memorizing partner-specific quirks, but instead support the emergence of communication protocols that scale with the number of agents.
    }
\label{fig:time}
\end{figure}

\paragraph{Training cost linearity.}
\autoref{fig:time} illustrates the relationship between population scaling and the total number of training iterations required for a communication group to reach the target in-group validation accuracy.
We evaluate $M = 16$ independently trained groups for each communication group size.
As detailed in \autoref{para_training_termination}, the number of training iterations is not predetermined, but is instead determined by an early stopping criterion based on a fixed benchmark in-group communication accuracy.
For both datasets, the relationship between communication group size ($N$) and the total number of training iterations is well described by a linear specification.
The Ramsey regression equation specification error test (RESET) shows no evidence of functional-form misspecification, with $p$-values of $0.730$ for MNIST and $0.954$ for CIFAR-10.
This approximate linearity is notable because the number of possible sender--receiver pairings in a fully connected communication group grows quadratically with $N$.
If agents were required to establish separate partner-specific conventions, i.e., idioglossia, the training burden would be expected to grow superlinearly with population size.
Instead, the observed linear trend suggests that larger populations do not solve communication by memorizing pairwise, partner-specific quirks.
Rather, the growing connectivity appears to induce scalable conventions, and, at least to a certain extent, universality, that allow individual agents to remain mutually compatible with an expanding set of partners at roughly constant per-agent training cost.

\begin{figure}[t]
    \centering
    \begin{subfigure}[b]{0.4\textwidth}
        \centering
        \includegraphics[width=\textwidth]{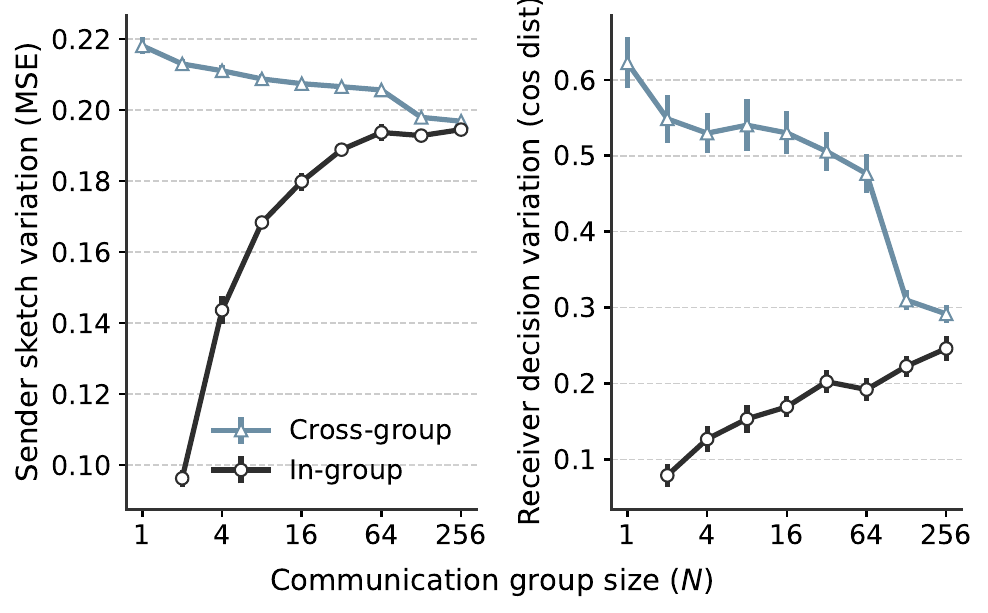}
        \caption{MNIST dataset}
    \end{subfigure}
    \hspace{0.05\textwidth}
    \begin{subfigure}[b]{0.4\textwidth}
        \centering
        \includegraphics[width=\textwidth]{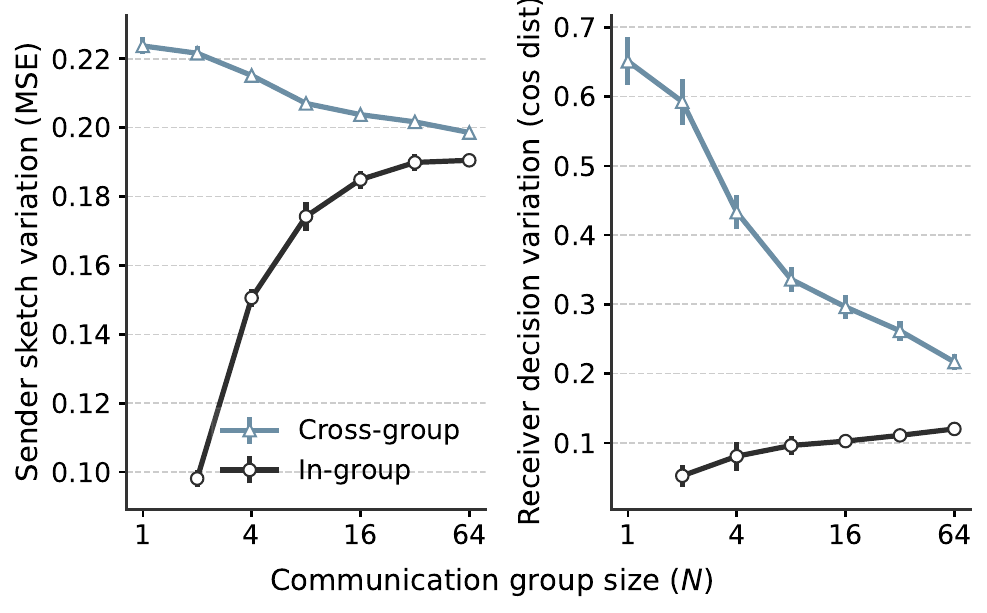}
        \caption{CIFAR-10 dataset}
    \end{subfigure}
\caption{
    Convergence of in-group and cross-group communicative variation across population scales.
    Sender-specific sketch variation (measured via MSE between sketches of the same target) and receiver-specific decision variation (measured via cosine distance between output logits) are evaluated for agent pairs drawn from the same group (in-group) and from independent groups (cross-group).
    Error bars represent 95\% confidence intervals.
    As we scale $N$, in-group variation rises while cross-group variation falls.
    %
    %
    This convergence illustrates that scaling social interactions prevents group-specific co-adaptation into a homogeneous idioglossia, instead utilizing in-group diversity as a regularizer to promote more universal, interoperable conventions.
    }
\label{fig:variation}
\end{figure}

\paragraph{Diversity toward universality.}
\autoref{fig:variation} further clarifies the mechanism that bridges the linear scaling behavior observed in \autoref{fig:time} with the increase of ZMI through population scaling observed in \autoref{fig:zmi}.
We measure sender-specific sketch variation and receiver-specific decision variation for pairs of agents drawn either from the same trained group or from independently trained groups.
Sender-specific sketch variation is measured as the mean squared error between a pair of created sketches drawn by the two different agents from the same target images.
Receiver-specific decision variation is measured as the cosine distance ($1 - \text{cosine similarity}$) between a pair of output decision logits;
here, for the cross-group receiver-specific decision variation, the input sketch is chosen at random from the two heterogeneous groups.
With increasing communication group size, in-group variation increases for both senders and receivers, showing that larger populations do not converge to a degenerate homogeneous protocol in which all agents behave identically.
In contrast, cross-group variation decreases with population size, indicating that independently trained populations become progressively less identifiable by their community memberships.
At the largest population scales, the gap between in-group and cross-group variation substantially diminishes.
This convergence indicates that agents in larger populations learn to accommodate a growing degree of communicative heterogeneity which prevents group-specific co-adaptation.
Simultaneously, this pressure to remain mutually compatible across diversifying partners acts as a structural regularizer that direct towards a certain type of universality that facilitates interoperability across disjoint groups.

\begin{figure}[t]
    \centering
    \begin{subfigure}[b]{0.36\textwidth}
        \centering
        \includegraphics[width=\textwidth]{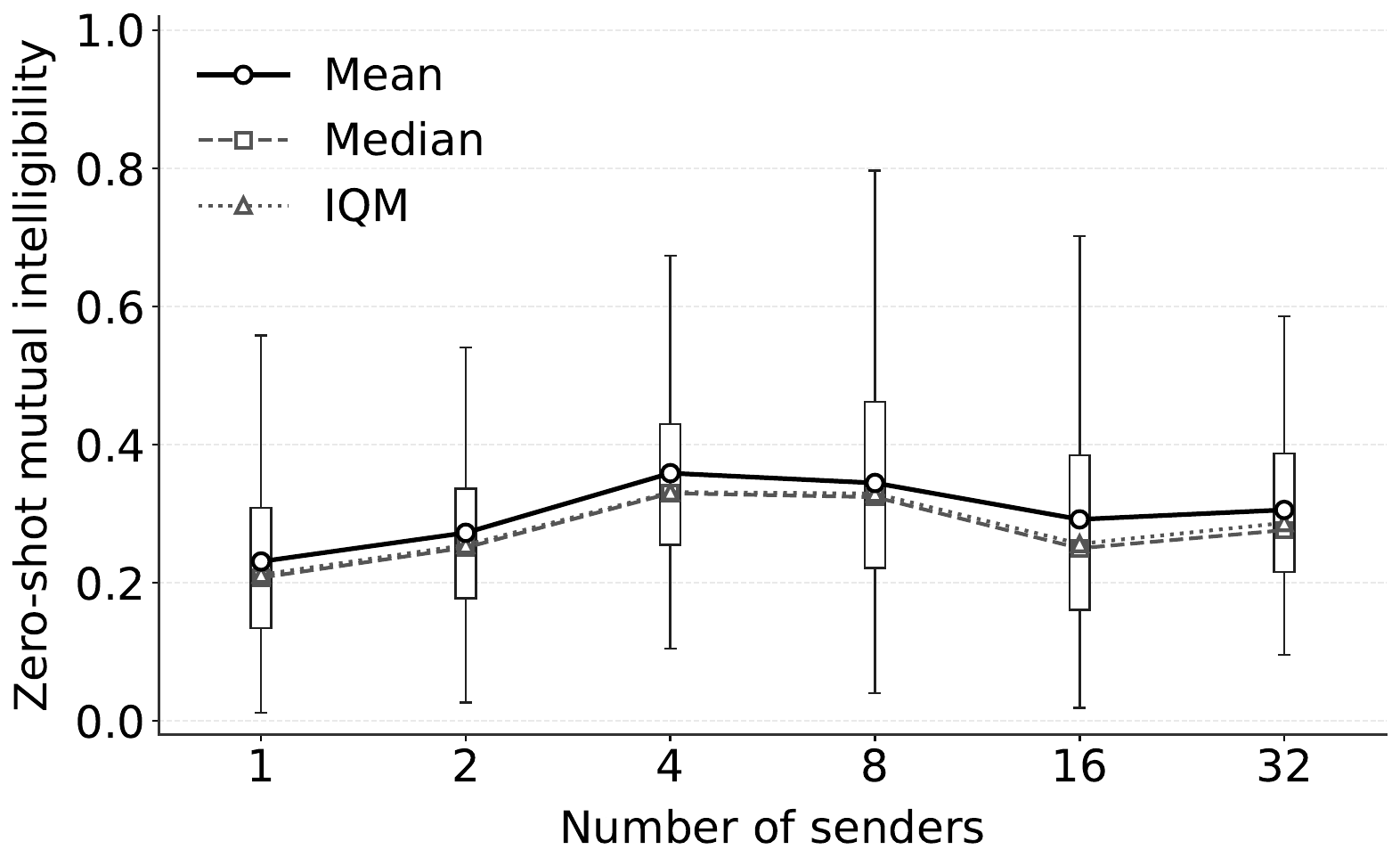}        
        \caption{MNIST: Multi-sender, solo receiver}
    \end{subfigure}
    \hspace{0.05\textwidth}
    \begin{subfigure}[b]{0.36\textwidth}
        \centering
        \includegraphics[width=\textwidth]{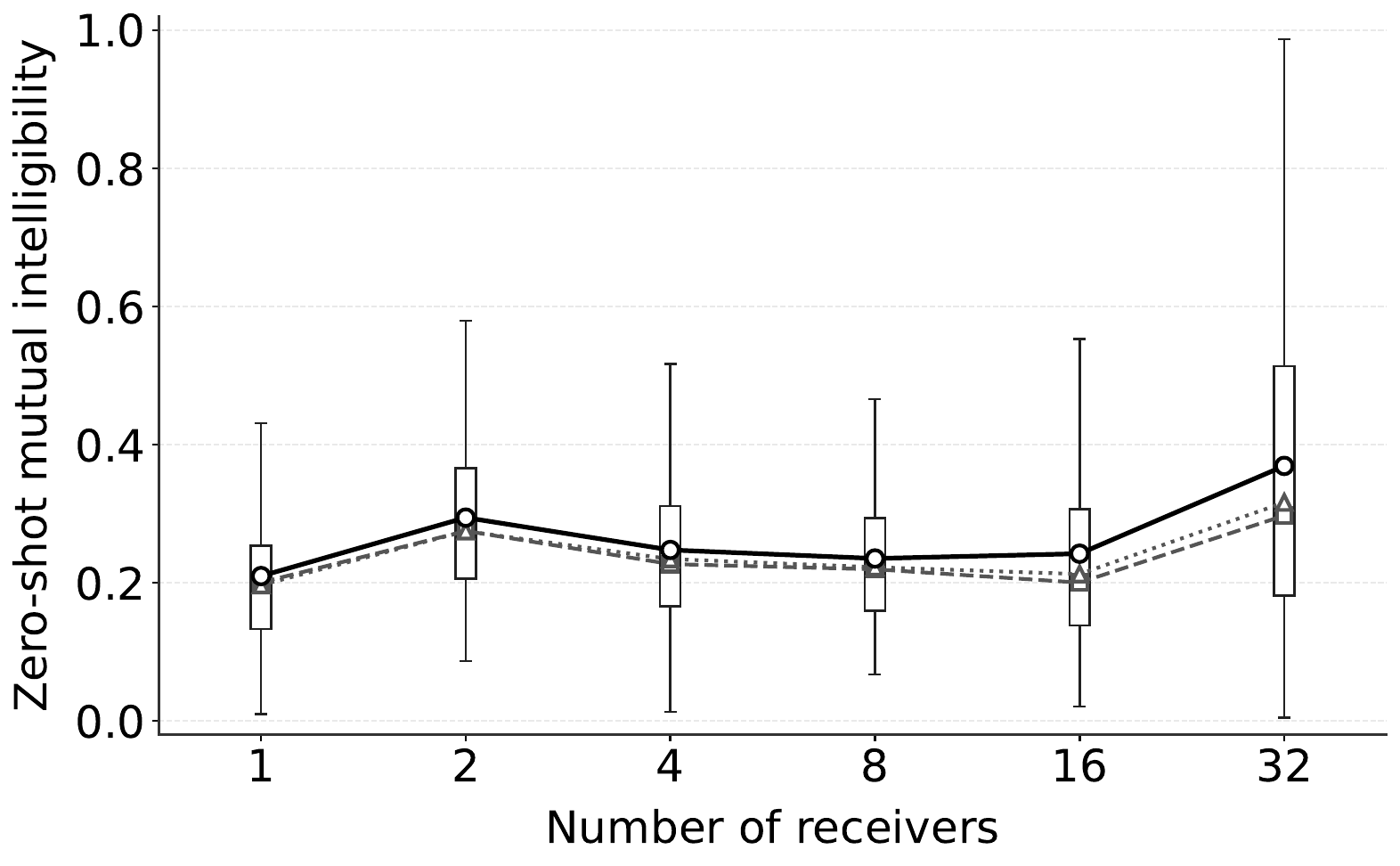}        
        \caption{MNIST: Multi-receiver, solo sender}
    \end{subfigure}
    \begin{subfigure}[b]{0.36\textwidth}
        \centering
        \includegraphics[width=\textwidth]{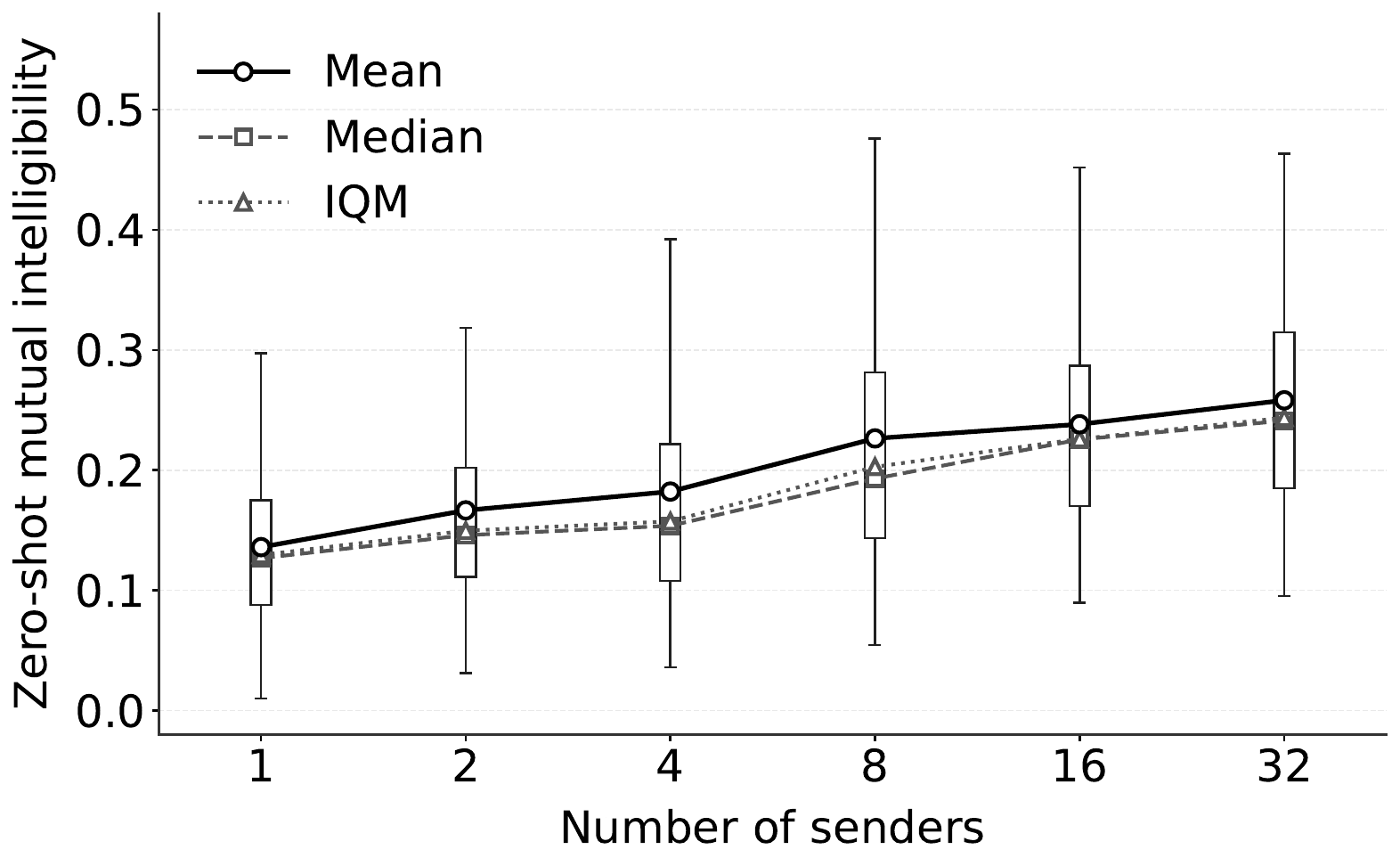}        
        \caption{CIFAR-10: Multi-sender, solo receiver}
    \end{subfigure}
    \hspace{0.05\textwidth}
    \begin{subfigure}[b]{0.36\textwidth}
        \centering
        \includegraphics[width=\textwidth]{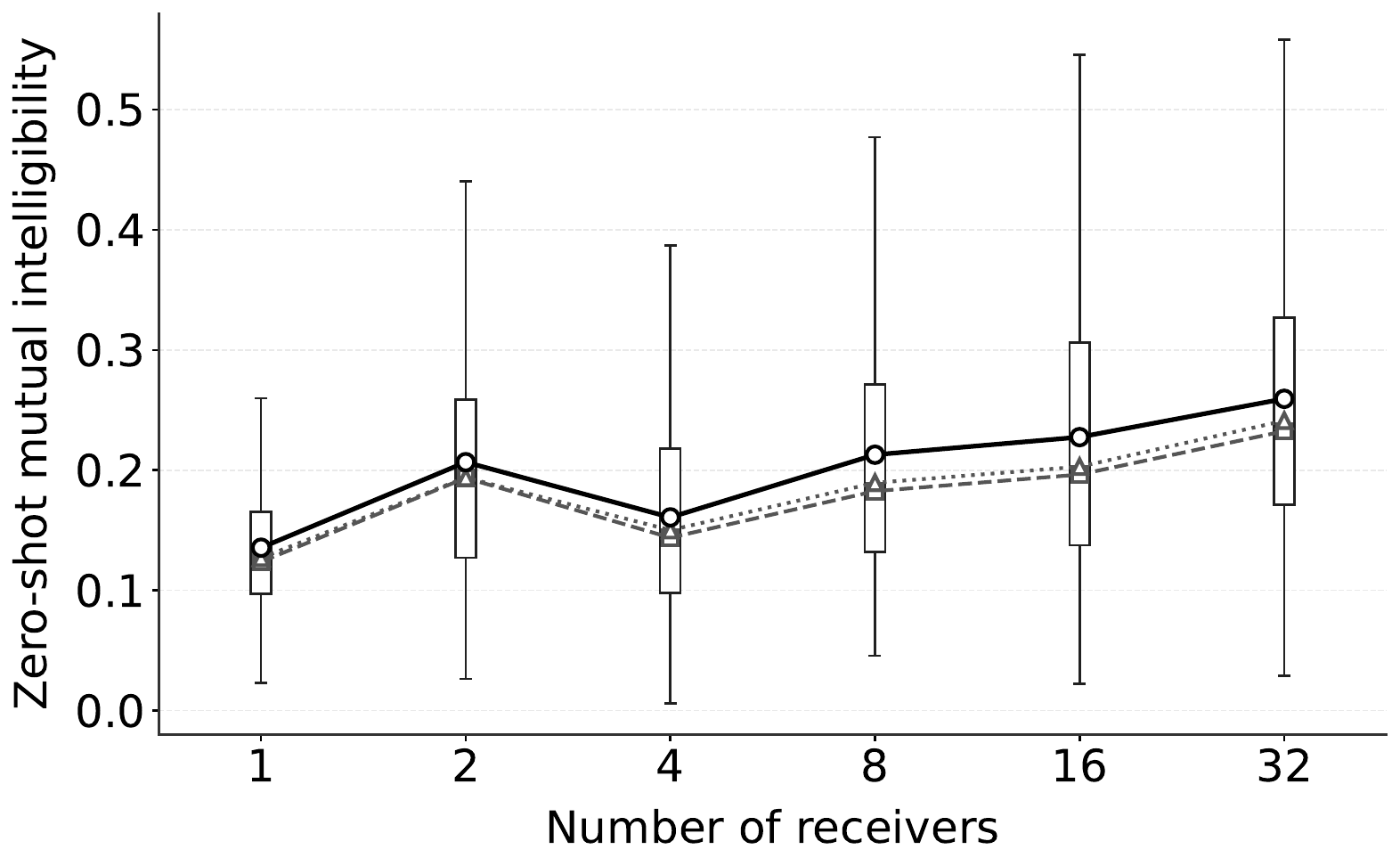}        
        \caption{CIFAR-10: Multi-receiver, solo sender}
    \end{subfigure}
\caption{
    Impact of centralized, one-to-many communication topologies on ZMI.
    ZMI is evaluated across increasing partner sizes on the MNIST ((a), (b)) and CIFAR-10 ((c), (d)) datasets. 
    Panels (a) and (c) depict topologies with a single receiver and $N$ senders, while panels (b) and (d) depict a single sender and $N$ receivers. 
    Although ZMI shows marginal improvements as $N$ grows, the rate of increase is substantially attenuated across all configurations compared to the fully connected setting (\autoref{fig:zmi}).
    This highlights that bidirectional diversity---varying both the generation and interpretation of signals—is necessary to effectively suppress partner-specific co-adaptation and drive the emergence of highly interoperable protocols.
    }
\label{fig:onetomany}
\end{figure}

\paragraph{The necessity of bidirectional diversity.}
To investigate the structural prerequisites for the high-fidelity ZMI observed in fully connected populations, we evaluate centralized, one-to-many communication graphs (\autoref{fig:onetomany}).
We isolate the scaling effect by training populations where a single receiver interacts with $N$ senders (panels (a), (c)) or a single sender interacts with $N$ receivers (panels (b), (d)).
While training these centralized topologies still requires linearly increasing iterations (\autoref{fig:onetomany_linearity} in the \autoref{app:linearity}), ZMI grows at a substantially slower pace across both datasets compared to the fully connected setting.
This discrepancy demonstrates that exposing only one side of the communication channel to a diverse population is insufficient to fully regularize against co-adaptation.
Rather, a static, centralized agent likely acts as a communicative anchor, inadvertently permitting the population to overfit to a single partner's specific idiosyncrasies rather than converging on a broadly shared convention.
These results suggest that robust social generalization is not merely a byproduct of encountering diverse partners but emerges from the dual, reciprocal cognitive demand of simultaneously interpreting heterogeneous signals and producing universally interpretable ones.

\begin{figure}[t]
    \centering
    \begin{subfigure}[b]{0.4\textwidth}
        \centering
        \includegraphics[width=\textwidth]{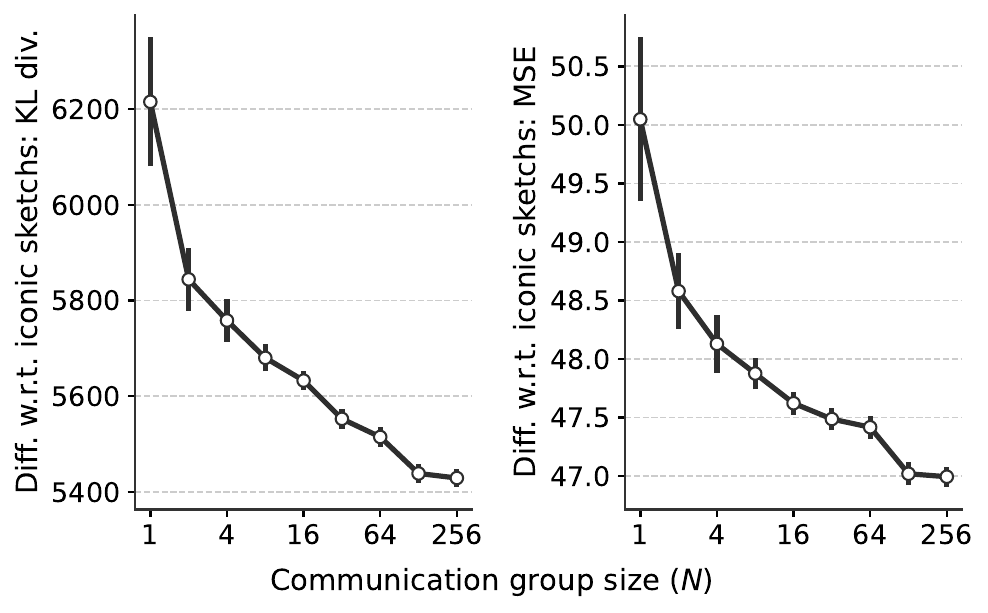}
        \caption{MNIST dataset}
    \end{subfigure}
    \hspace{0.05\textwidth}
    \begin{subfigure}[b]{0.4\textwidth}
        \centering
        \includegraphics[width=\textwidth]{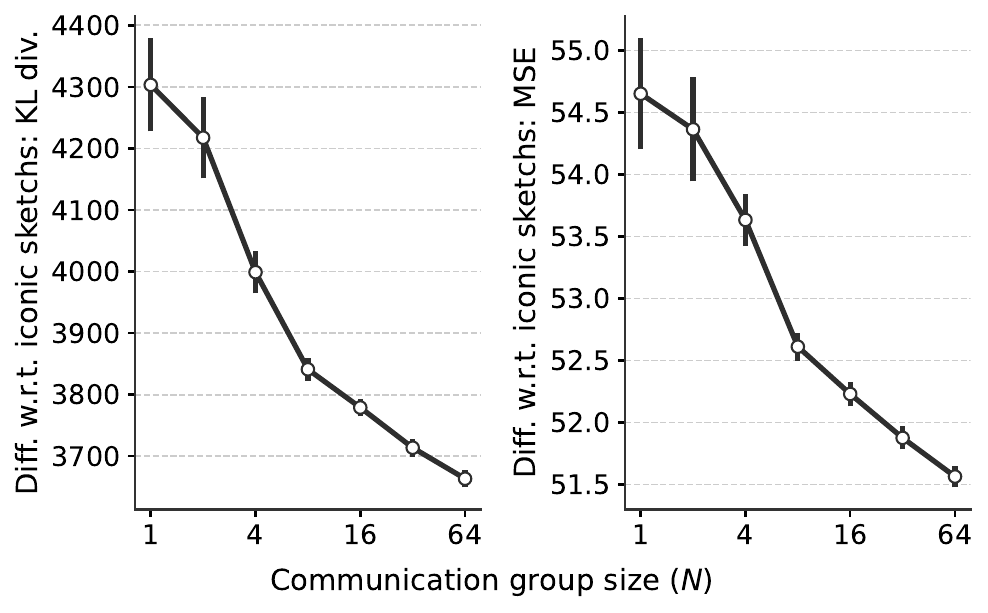}
        \caption{CIFAR-10 dataset}
    \end{subfigure}
\caption{
    Effect of population scaling on the perceptual grounding of emergent sketches.
    The visual similarity between agent-generated sketches and reference ``iconic sketches'' (optimized purely for perceptual similarity) is evaluated across varying communication group sizes ($N$) for the MNIST (a) and CIFAR-10 (b) datasets.
    Similarity is measured using KL divergence in a VAE latent space (left panels) and pixel-level mean squared error (right panels). 
    Error bars represent the 95\% confidence interval.
    For both datasets and both metrics, divergence from the iconic reference systematically decreases as population size grows, indicating that larger groups favor perceptually grounded communication over arbitrary protocols.
    }
\label{fig:iconic}
\end{figure}
\begin{figure}[t]
    \centering
    \begin{subfigure}[b]{0.4\textwidth}
        \centering
        \includegraphics[width=\textwidth]{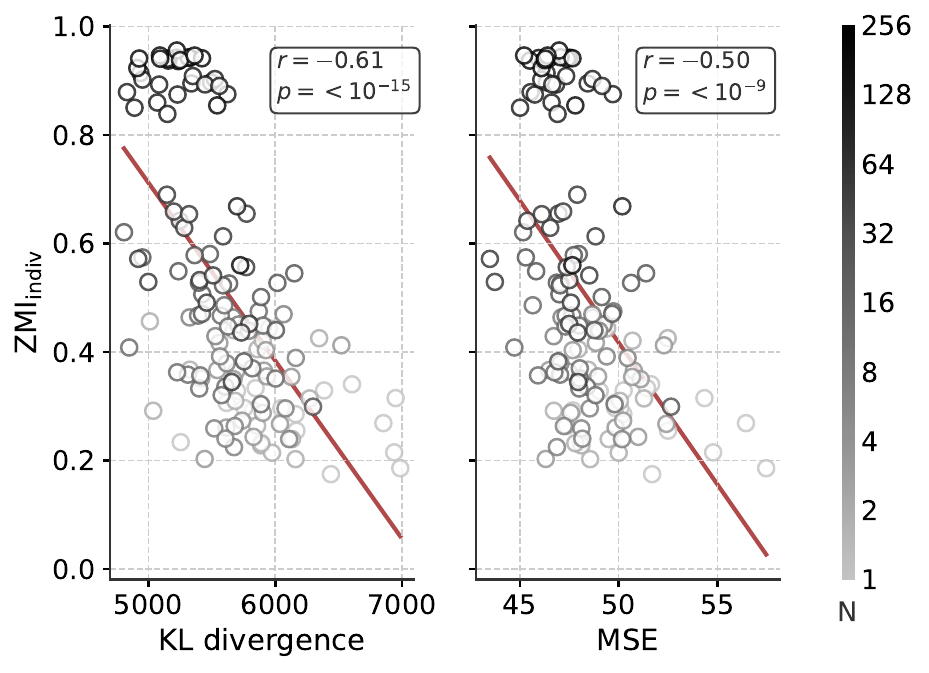}
        \caption{MNIST dataset}
    \end{subfigure}
    \hspace{0.05\textwidth}
    \begin{subfigure}[b]{0.4\textwidth}
        \centering
        \includegraphics[width=\textwidth]{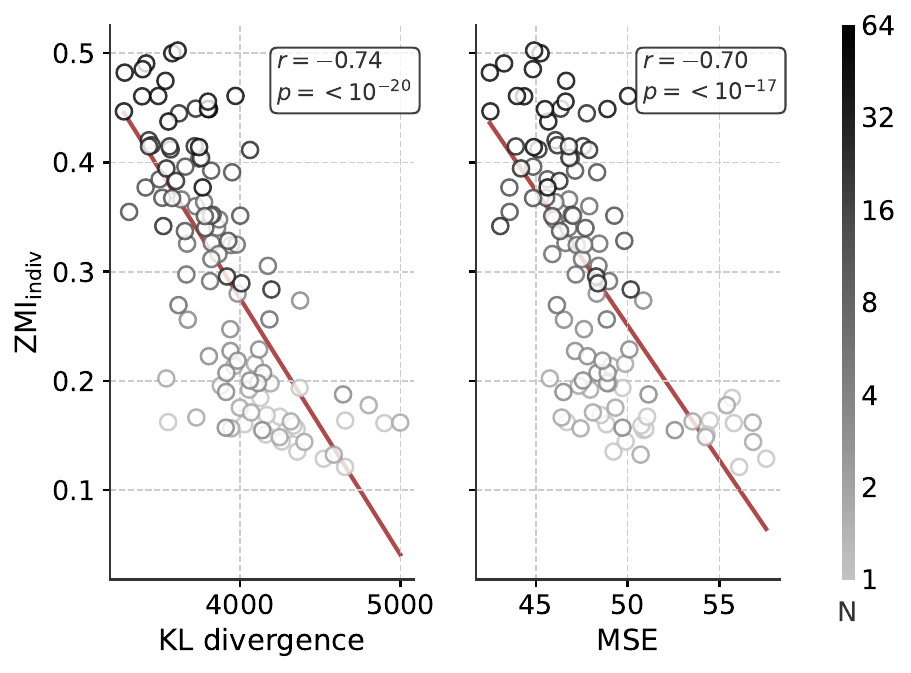}
        \caption{CIFAR-10 dataset}
    \end{subfigure}
\caption{
    Relationship between perceptual grounding and $\text{ZMI}_{\text{indiv}}$.
    Scatterplots display the individual agent performance across different population sizes, evaluated on MNIST (a) and CIFAR-10 (b).
    The x-axes represent the divergence from iconic sketches (KL divergence and MSE), and the y-axes represent the individual ZMI.
    Dots are color-coded by the population size ($N$) of the agent's training group.
    Linear regression fits (solid lines) demonstrate negative correlations across all configurations.
    This confirms that greater perceptual grounding is intrinsically associated with higher communicative success with unseen strangers.\vspace{-2mm}
    }
\label{fig:iconic_indiv}
\end{figure}
\subsection{Perceptual grounding as a mechanism for universality}
\label{sec:perceptual}
So far, we have established that population scaling prevents partner-specific co-adaptation and drives populations toward a broadly interoperable protocol.
We hypothesize that the specific nature of this emerging universality is rooted in the modality itself: visually grounded communication.
Without the capacity to rely on arbitrary, agreed-upon idioglossia, agents in large populations may increasingly anchor their communicative protocols to the objective, perceptual features of the target images.
To quantitatively evaluate this hypothesis, we first construct a set of reference ``iconic sketches'' for the target images.
Following \citet{mihai2021learning}, these reference sketches are generated by minimizing a perceptual loss against the target images, without any communicative objective.
Refer to \autoref{app:perceptual_sketches} for details.
We then measure the similarity between the emergent sketches produced by the communicative agents and the perceptually grounded sketches using two metrics: mean squared error (MSE) to capture pixel-level structural differences, and Kullback-Leibler (KL) divergence computed in the latent space of a variational autoencoder (VAE) to capture higher-level semantic deviations.
As illustrated in \autoref{fig:iconic}, population scaling systematically increases the visual similarity between the emergent protocols and the iconic references.
Across both datasets, we observe a consistent, monotonic decrease in both KL divergence and MSE as the communication group size ($N$) increases.
This trend demonstrates that multi-agent optimization pressure increasingly favors protocols built on features that structurally resemble the visual inputs.
As agents are exposed to a wider variety of partners, their sketches become measurably more grounded in the perceptual reality of the environment.

Furthermore, we observe that this shift toward perceptual grounding is directly linked to the capability to communicate with strangers.
\autoref{fig:iconic_indiv} presents the relationship between the visual similarity metrics of individual agents and their corresponding individual zero-shot mutual intelligibility ($\text{ZMI}_{\text{indiv}}$).
The results demonstrate a strong negative correlation between the visual divergence metrics and ZMI across all settings, with Pearson correlation ranging from $-0.50$ to $-0.74$, all $p < 10^{-9}$.
This tight correlation suggests that the increase in ZMI is not merely a coincidental byproduct of population size, but is fundamentally mediated by the degree to which agents adopt perceptually grounded signaling strategies.
It is important to note, however, that these emergent protocols do not achieve photographic realism.
While the metrics indicate a robust relative shift toward perceptual grounding, qualitative inspection of the emergent sketches (provided in \autoref{app:examples}) reveals that they remain highly abstract to the human eye, even in the largest population settings.
Therefore, we do not claim that large populations inherently solve the task by drawing perfect replicas of the targets.
Rather, the data suggests that scaling the social environment systematically biases the optimization trajectory away from arbitrary signs and toward shared perceptual features, providing a sufficient anchor to establish mutual intelligibility among strangers.
\section{Related work}

\paragraph{Emergent communication and generalization.}
Emergent communication studies how protocols arise between autonomous agents through interaction and task rewards \citep{foerster2016learning,sukhbaatar2016learning,lazaridou2017multi,havrylov2017emergence}.
Early work showed that deep RL agents can invent signaling systems for referential and coordination games \citep{jorge2016learning}, but subsequent studies revealed a recurring pathology: emergent protocols often become partner-specific and fail to generalize, lacking structural properties associated with human language such as systematicity and compositionality \citep{kottur2017natural,chaabouni2020compositionality}.
As a result, most work evaluates generalization either at the input level, measuring robustness to novel stimuli or distribution shift \citep{mu2021emergent,rita2022emergent}, or at the language level, measuring compositional or systematic structure under explicit pressures \citep{hill2020environmental,carmeli2024concept,mordatch2018emergence}.
In contrast, we focus on social generalization: whether agents trained in strictly disconnected communities can interpret signals from strangers.\vspace{-1mm}

\paragraph{Zero-shot coordination, cross-play, and convention symmetries.}
Partner generalization in multi-agent RL is formalized as Zero-Shot Coordination (ZSC), where agents must coordinate with independently trained partners \citep{hu2020other}.
A standard evaluation protocol is cross-play (XP), which measures performance when pairing independently trained policies rather than self-play.
Approaches such as other-play and follow-ups improve cross-play by discouraging brittle partner-specific conventions, for example by enforcing invariances or reasoning over partner uncertainty \citep{hu2020other,hu2021off,treutlein2021new,carroll2019utility}.
Communication introduces an additional difficulty: symbolic channels admit near-permutation invariance over tokens, meaning many equally valid equilibria can emerge that are mutually unintelligible across independently trained populations.
We adapt the XP evaluation protocol to the semantic layer and term this target zero-shot mutual Intelligibility (ZMI), defined as whether independently learned signal meanings align across populations rather than whether policies simply coordinate behavior.\vspace{-1mm}

\paragraph{Human--agent communication and alignment.}
A related line studies emergent communication when agents must align with humans or natural language.
\citet{lazaridou2020multiagent} show how multi-agent interaction can help connect emergent protocols with natural language and emphasize generalization beyond closed populations toward communication in the wild.
Complementary work demonstrates that agents can acquire behaviors that collaborate with humans even without paired human data \citep{strouse2021collaborating}.
These works provide an important precedent for evaluating generalization across disjoint partner distributions, but they typically rely on external structure such as natural language or human evaluation signals.
Our goal is to isolate the agent-to-agent interoperability problem and study which training mechanisms drive semantic convergence when both sides are independently trained artificial populations.\vspace{-1mm}

\paragraph{Language drift, cultural evolution, and iterated learning.}
The stability and transfer of communication conventions has long been studied in cultural evolution.
Iterated learning models show how transmission bottlenecks and population turnover regularize languages toward structure that is easier to learn and transmit \citep{kirby2001spontaneous,kirby2014iterated,smith2003iterated}.
In modern deep emergent communication, related failures appear as language drift, where co-adaptation between agents leads to protocol divergence from externally interpretable semantics.
Several approaches mitigate drift by shaping transmission or grounding, including seeded iterated learning \citep{lu2020countering} and perceptual grounding constraints \citep{lee2019countering}.
From this perspective, interaction topology can be viewed as an explicit transmission mechanism that shapes which conventions propagate and how quickly they stabilize.
We study this effect through cross-population semantic interoperability, rather than through compositionality or external interpretability alone.\vspace{-1mm}

\paragraph{Population size and structure in human socio-linguistics.}
Human socio-linguistic and cognitive research shows that population size, social connectivity, and input diversity jointly shape communication system structure. Large, socially heterogeneous populations tend to favor more regular, systematic, and compositional linguistic systems, whereas small, tightly knit communities often maintain more irregular or idiosyncratic structure \citep{lupyan2010language,trudgill2011sociolinguistic,wray2007social}. Importantly, population-scale effects cannot be reduced to input variability alone: demographic size predicts linguistic simplification even when variability is controlled \citep{atkinson2015speaker}. At the cognitive level, exposure to variable inputs improves detection of invariant structure and promotes abstraction across surface variation \citep{gomez2002variability,perry2010learn}. Social network size further influences communicative success, with larger and more diverse networks associated with stronger semantic coordination abilities \citep{lev-ari2016social}. Experimental cultural evolution studies likewise show that increasing community size promotes more systematic and compressible signaling systems \citep{raviv2019larger}. Together, these findings suggest that population scale acts as a regularizer on communication systems, discouraging partner-specific conventions and promoting broadly interpretable structure. Our results in emergent multi-agent communication parallel these observations: increasing population scale improves cross-group intelligibility and interoperability.\vspace{-1mm}

\paragraph{Population scaling in multi-agent communication.}
Recent work investigates how population size and diversity shape emergent communication, demonstrating that exposure to varied partners qualitatively alters protocol structures \citep{graesser2019emergent,kim2021emergent,chaabouni2022emergent,michelrevisiting}. Parallel findings in multi-agent learning highlight that population-based training is critical for policy robustness and zero-shot generalization \citep{jaderberg2017population,strouse2021collaborating,baker2019emergent}. Our research synthesizes and extends these lines of inquiry in two key ways: first, by establishing zero-shot mutual intelligibility (ZMI) across disjoint populations as the primary objective; and second, by demonstrating that the regularizing effects of large populations—which actively suppress idioglossia—can be achieved at a linear training cost. Ultimately, this indicates that the necessity to accommodate in-group diversity, rather than dense pairwise memorization, fundamentally drives the emergence of universal interoperability.\vspace{-1mm}

\paragraph{Visually grounded and sketch-based communication.}
Discrete-symbol protocols are purely conventional and therefore vulnerable to symmetry and idiosyncratic equilibria across independent training runs.
Visually grounded signaling can stabilize semantics by anchoring messages in perceptual structure.
Emergent sketching enables agents to communicate via structured drawings with iconic properties \citep{mihai2021learning}, building on neural representations for sketch data \citep{ha2018neural} and interactive sketch-based communication tasks \citep{lian2024interactive}.
Unlike multimodal settings that use symbols to describe images, sketching uses images directly as signals.
This provides a shared perceptual bias that reduces the space of equally valid communication equilibria and facilitates semantic alignment across disjoint populations.\vspace{-1mm}


\paragraph{Summary.}
We formalize ZMI as a distinct axis of social generalization, evaluating the communicative success between strictly disjoint, independently trained populations. Empirically, we demonstrate that high-fidelity ZMI is driven by the critical synergy of two factors: large-scale, bidirectional partner exposure, which acts as a structural regularizer against brittle idioglossia, and a visually grounded signaling modality, which anchors the resulting universal conventions in objective perceptual reality.\vspace{-1mm}
\section{Discussion and Conclusion}

We formalized zero-shot mutual intelligibility (ZMI) as a distinct and measurable axis of generalization in emergent communication.
By leveraging emergent sketching as a visually grounded modality, we demonstrated that high-fidelity communication can be achieved between strictly disjoint, independently trained artificial populations.
Crucially, our findings reveal that scaling the social environment---specifically through fully connected, bidirectional interactions---acts as a powerful structural regularizer.
Rather than collapsing into partner-specific idioglossia, larger populations accommodate a baseline of communicative heterogeneity. 
This internal diversity compels agents to converge upon universal interoperability anchored in the objective visual perceptuality.
The linear scaling of training costs in fully connected groups suggests that agents are not merely memorizing a set of local rules, but are fundamentally adopting a more generalized, robust communicative strategy.
This shifts the focus of emergent communication from optimizing closed-system efficiency toward designing open-ended protocols capable of seamlessly integrating with unseen partners.

Our current framework has notable limitations. First, while the emergent conventions become increasingly grounded in the visual features of the target images, the resulting sketches remain highly abstract to human observers. The current scaling regime is sufficient for agent-to-agent universality, but achieving true human-interpretable, iconic communication likely requires additional mechanisms. Furthermore, our empirical analysis relies on static, fully connected communication graphs. Scaling to significantly larger populations will eventually render fully connected topologies computationally prohibitive, necessitating the exploration of sparse, dynamic, or spatially structured social networks for achieving sublinear training cost.
%
%
\clearpage


\clearpage
\bibliography{main}
\bibliographystyle{tmlr}

\clearpage
\appendix
\section{Additional Results on CelebA}
\label{app:celebA}

To further examine the robustness of our findings, we evaluate the relationship between population scaling and zero-shot mutual intelligibility (ZMI) on an additional dataset, CelebA~\citep{liu2015faceattributes}. While MNIST and CIFAR-10 provide compact benchmarks based on handwritten digits and natural object categories, respectively, CelebA consists of face images annotated with multiple semantic attributes. This provides a complementary evaluation setting with substantially more images and a visually distinct domain.

For the CelebA dataset, we downsample images to $32 \times 32$ resolution. We use the official training split for training and the official validation split for evaluation; the official test split is not used. The dataset statistics used in our experiments are summarized in \autoref{tab:dataset_stats}.
The reference in-group validation communication accuracy is set to 0.95 for the early stopping.

For MNIST and CIFAR-10, we use the target-variation protocol described in the main text: the sender and receiver observe different target images that belong to the same semantic category. Thus, successful communication requires agents to coordinate at the class or concept level, rather than by matching identical visual instances.
For CelebA, we instead abstain from using target variation and provide the sender and receiver with identical target images. We adopt this choice to further diversify the experimental setup, testing whether the population-scaling trend in ZMI also appears under an instance-level, conventional referential game rather than a semantic target-variation setting.

\begin{figure}[t]
    \centering
    \includegraphics[width=0.9\textwidth]{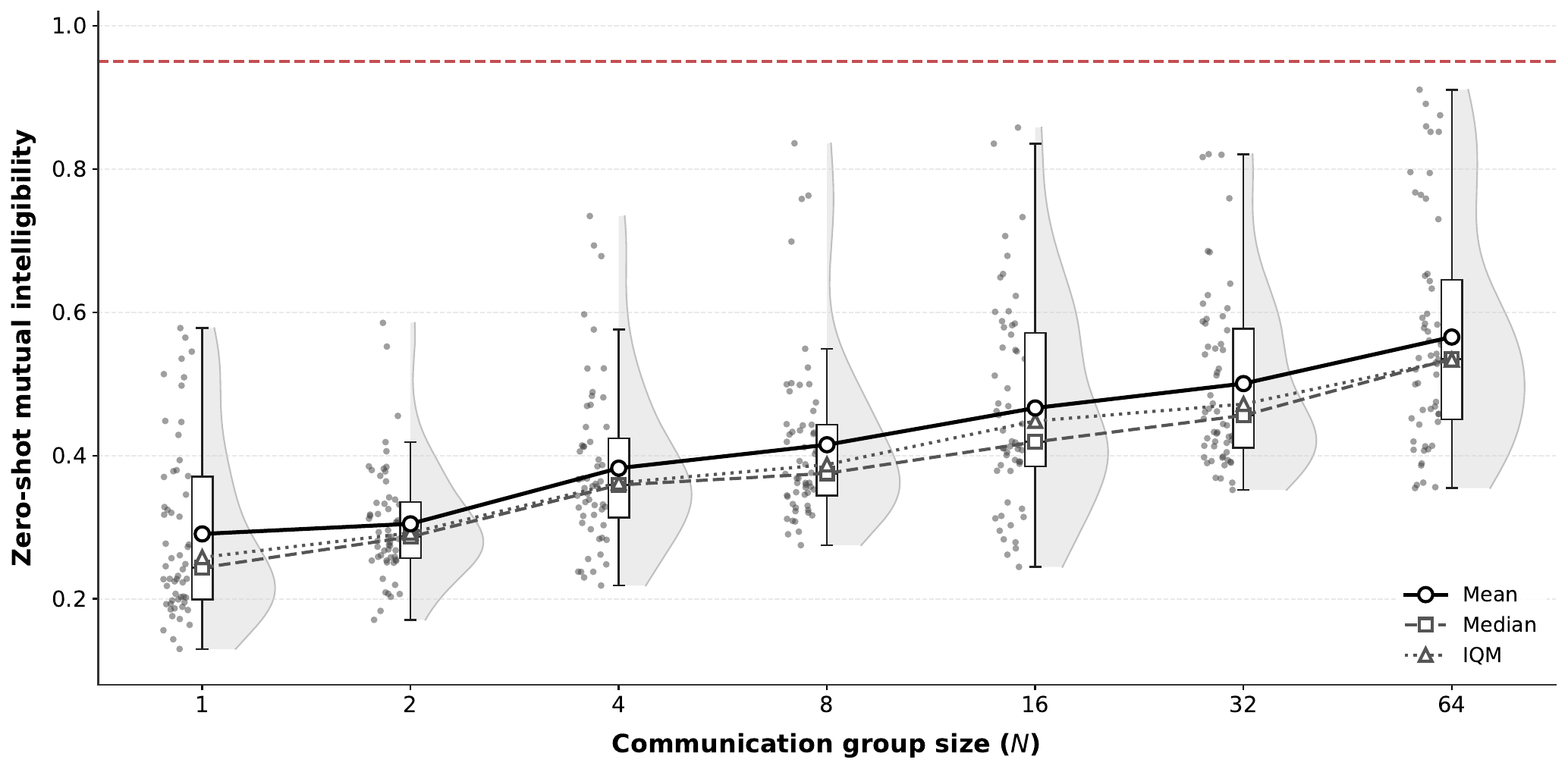}        
    \caption{
        Zero-shot mutual intelligibility (ZMI) evaluated on CelebA dataset.
        Each dot represents group-level ZMI (\autoref{group-zmi}).
        The number of groups $M$ is set to 8, thereby the number of dots is 56 ($8 \times 7$) for each population scale.
        The solid line represents the population-level ZMI (\autoref{population-zmi}), along with median (dashed) and interquartile mean (IQM; dotted).
        As for the datasets described in the main text, ZMI evaluated on the CelebA also increases substantially by scaling the communication group size ($N$).
        The in-group communication accuracy, which is represented as a red, dashed horizontal line, is set to 0.95.
        }
    \label{fig:zmi_celebA}
\end{figure}

As shown in \autoref{fig:zmi_celebA}, increasing the population size again leads to a consistent improvement in ZMI scores. This trend mirrors the results observed on MNIST and CIFAR-10 in the main text, providing additional evidence that the relationship between population scaling and zero-shot mutual intelligibility is not specific to digit or object classification benchmarks, but also extends to a visually richer face-attribute dataset.
\clearpage
\begin{table}[t]
    \centering
    \renewcommand{\arraystretch}{1.5}
    \setlength{\tabcolsep}{5pt}
    \begin{tabular}{lccrrl}
        \toprule
        Dataset & Resolution & Channels & Train & Eval. & Labels \\
        \midrule
        MNIST~\citep{lecun1998gradient}
        & $28 \times 28$ & $1$ & $60{,}000$ & $10{,}000$
        & $10$ digit classes \\

        CIFAR-10~\citep{krizhevsky2009cifar}
        & $32 \times 32$ & $3$ & $50{,}000$ & $10{,}000$
        & $10$ object classes \\

        CelebA~\citep{liu2015faceattributes}
        & $32 \times 32$ & $3$ & $162{,}770$ & $19{,}867$
        & $40$ binary facial attributes \\
        \bottomrule
    \end{tabular}
    \caption{
    Dataset statistics used in our experiments. For CelebA, images are downsampled to $32 \times 32$, and the official validation split is used for evaluation; the official test split is not used.
    }
    \label{tab:dataset_stats}
\end{table}

\clearpage
\section{Robustness Analyses}
\label{app:robustness}
In this section, we conduct a series of experiments to evaluate the robustness of the positive relationship between population scaling and the increase in ZMI.
Specifically, for both the MNIST and CIFAR-10 datasets, we first vary the reference in-group validation accuracy by increasing and decreasing its value.
We then vary the number of strokes in the sketches produced from the sender images.
Finally, we examine the effect of reducing the number of candidate images available to the receiver agents.
Note that, in the main text, the number of candidate images is fixed at 10, corresponding to the number of classes in both datasets.
Since this already exhausts the available classes, increasing the number of candidate images beyond 10 is not feasible, and we therefore only consider smaller candidate sets.
Also, because of the time constraint, for these additional experiments, we set the largest communication group size to $N = 32$.
\subsection{Varying the reference in-group validation accuracy}
\begin{figure}[ht]
    \centering
    \begin{subfigure}[b]{\textwidth}
        \centering
        \includegraphics[width=0.8\textwidth]{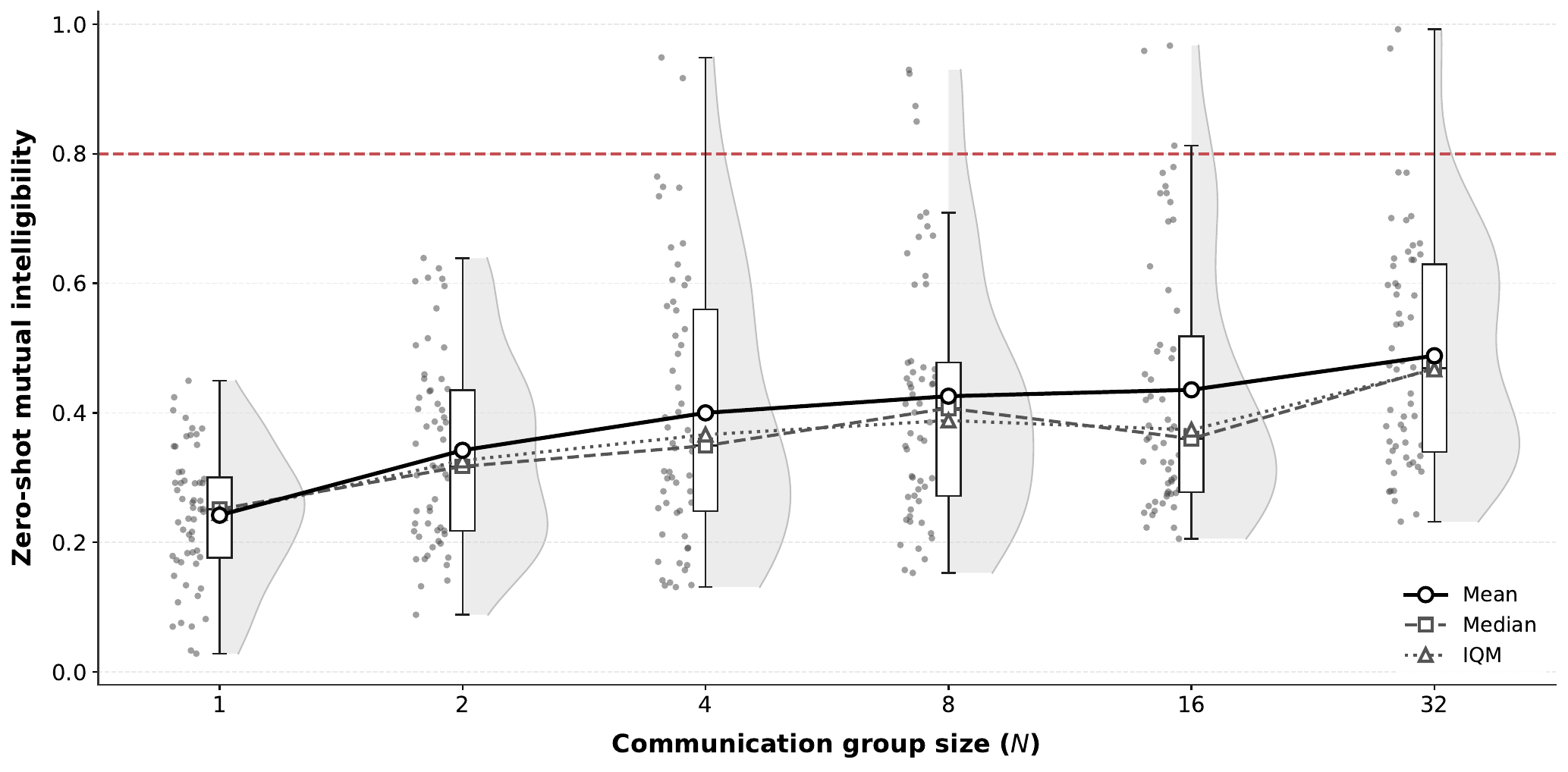}        
        \caption{MNIST dataset: Ref. val. acc. set to 0.80}
    \end{subfigure}
    \hfill
    \begin{subfigure}[b]{\textwidth}
        \centering
        \includegraphics[width=0.8\textwidth]{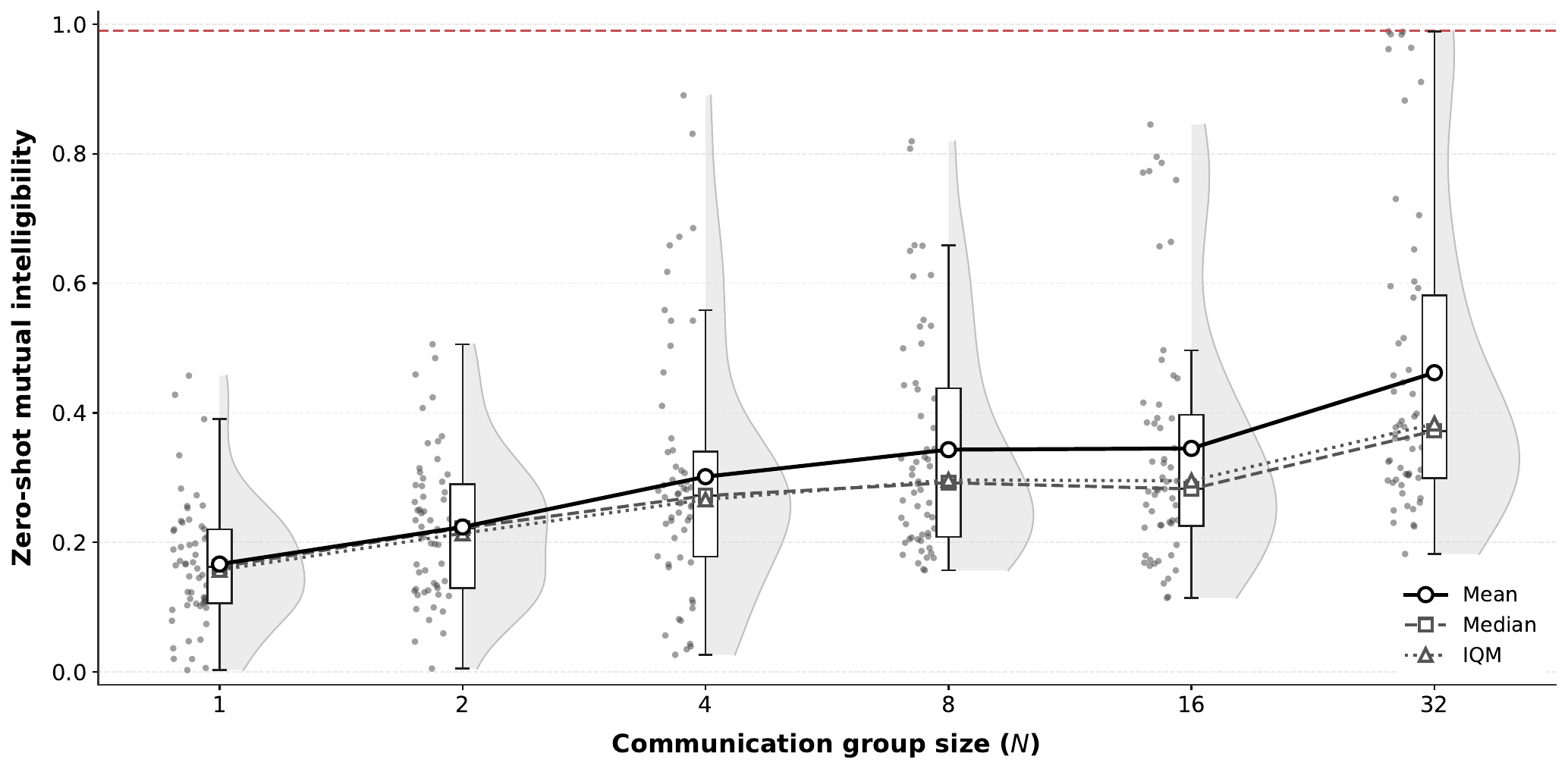}        
        \caption{MNIST dataset: Ref. val. acc. set to 0.99}
    \end{subfigure}
\caption{
    ZMI evaluated on the MNIST dataset under varying reference in-group validation communication accuracies.
    In the main text, the reference accuracy is set to 0.95.
    In the top panel (a), the reference accuracy is decreased to 0.8, whereas in the bottom panel (b), it is increased to 0.99.
    We observe that, for the MNIST dataset, the increase in ZMI with communication group size remains robust under both lower and higher reference in-group validation accuracies.
    }
\label{fig:zmi_ref_mnist}
\end{figure}

\begin{figure}[ht]
    \centering
    \begin{subfigure}[b]{\textwidth}
        \centering
        \includegraphics[width=0.8\textwidth]{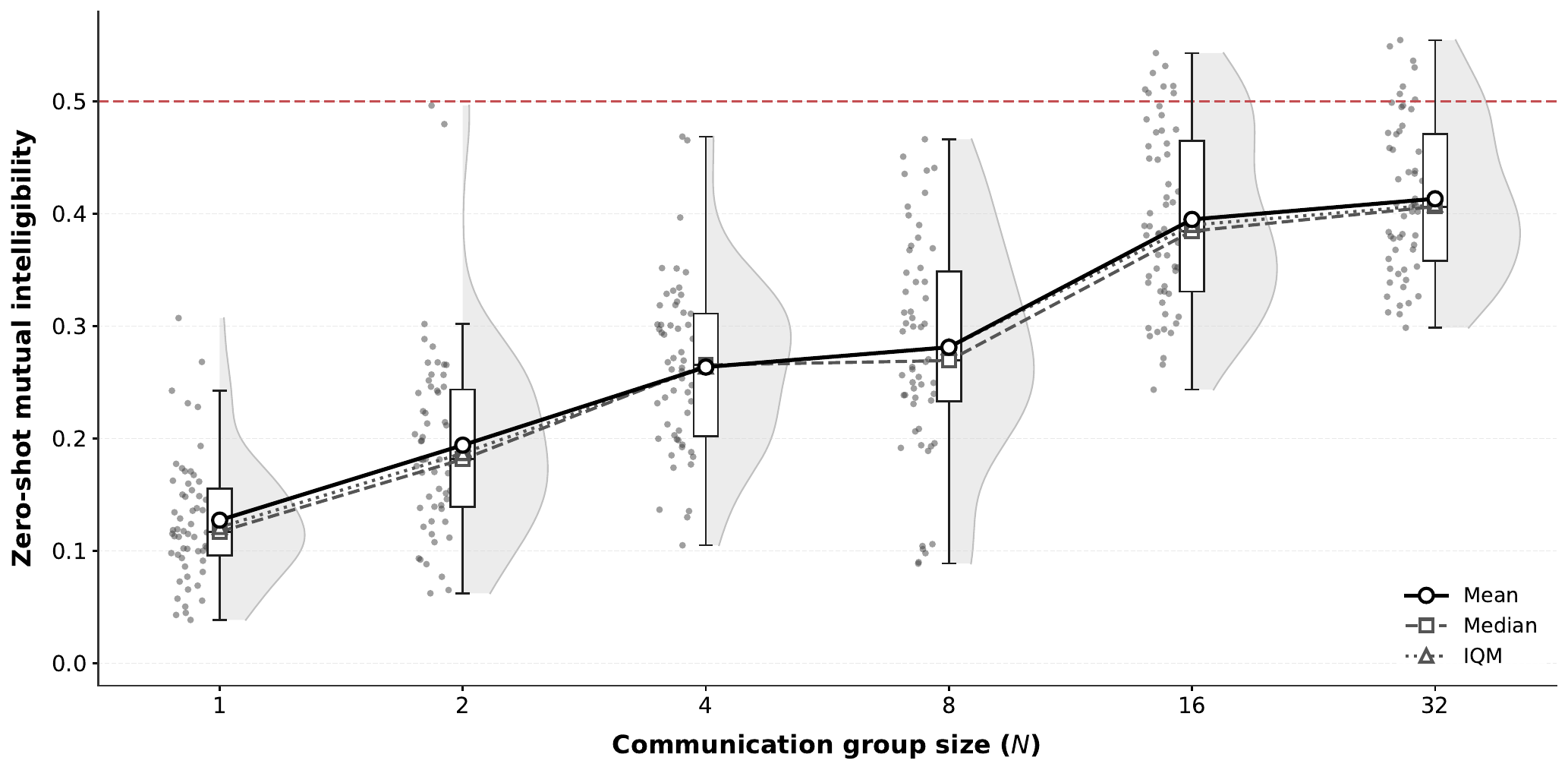}
        \caption{CIFAR-10 dataset: Ref. val. acc. set to 0.50}
    \end{subfigure}
    \hfill
    \begin{subfigure}[b]{\textwidth}
        \centering
        \includegraphics[width=0.8\textwidth]{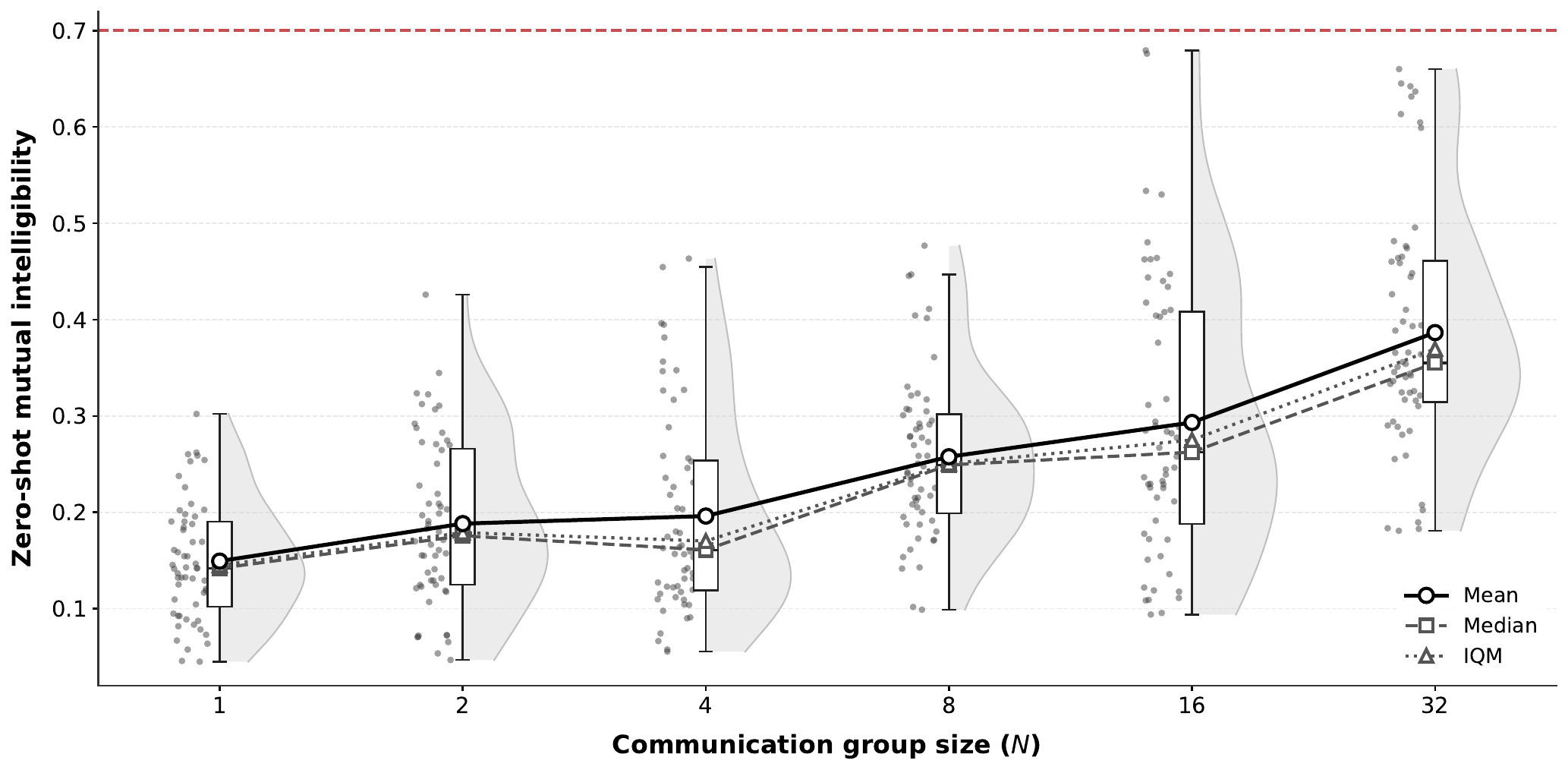}
        \caption{CIFAR-10 dataset: Ref. val. acc. set to 0.70}
    \end{subfigure}
\caption{
    ZMI evaluated on the CIFAR-10 dataset under varying reference in-group validation communication accuracies.
    In the main text, the reference accuracy is set to 0.6.
    In the top panel (a), the reference accuracy is decreased to 0.5, whereas in the bottom panel (b), it is increased to 0.7.
    We observe that, for the CIFAR-10 dataset, the increase in ZMI with communication group size remains robust under both lower and higher reference in-group validation accuracies.
    }
\label{fig:zmi_ref_cifar}
\end{figure}

The computation of ZMI depends on the choice of a reference in-group validation communication accuracy.
In the main experiments, this reference value is fixed to $0.95$ for MNIST and $0.60$ for CIFAR-10.
To examine whether the observed relationship between population scaling and ZMI is sensitive to this choice, we repeat the analysis after perturbing the reference accuracy upward and downward for each dataset.
For MNIST, we evaluate reference accuracies of $0.80$ and $0.99$.
For CIFAR-10,  we evaluate reference accuracies of $0.50$ and $0.70$.
\autoref{fig:zmi_ref_mnist} shows the results for MNIST.
Although changing the reference accuracy affects the absolute scale of the measured ZMI, the qualitative trend remains unchanged.
In both the lower-reference and higher-reference settings, ZMI increases as the communication group size grows.
This indicates that the positive relationship between population size and ZMI on MNIST is not an artifact of the particular reference accuracy used in the main text.
\autoref{fig:zmi_ref_cifar} shows the corresponding results for CIFAR-10.
Again, the same qualitative pattern is preserved under both lower and higher reference accuracies.
Together, these results support the conclusion that the increase in ZMI with larger communication populations does not depend critically on the precise reference accuracy used for terminating the training process.
\clearpage
\subsection{Varying the number of strokes in sketches}
\begin{figure}[ht]
    \centering
    \begin{subfigure}[b]{\textwidth}
        \centering
        \includegraphics[width=0.8\textwidth]{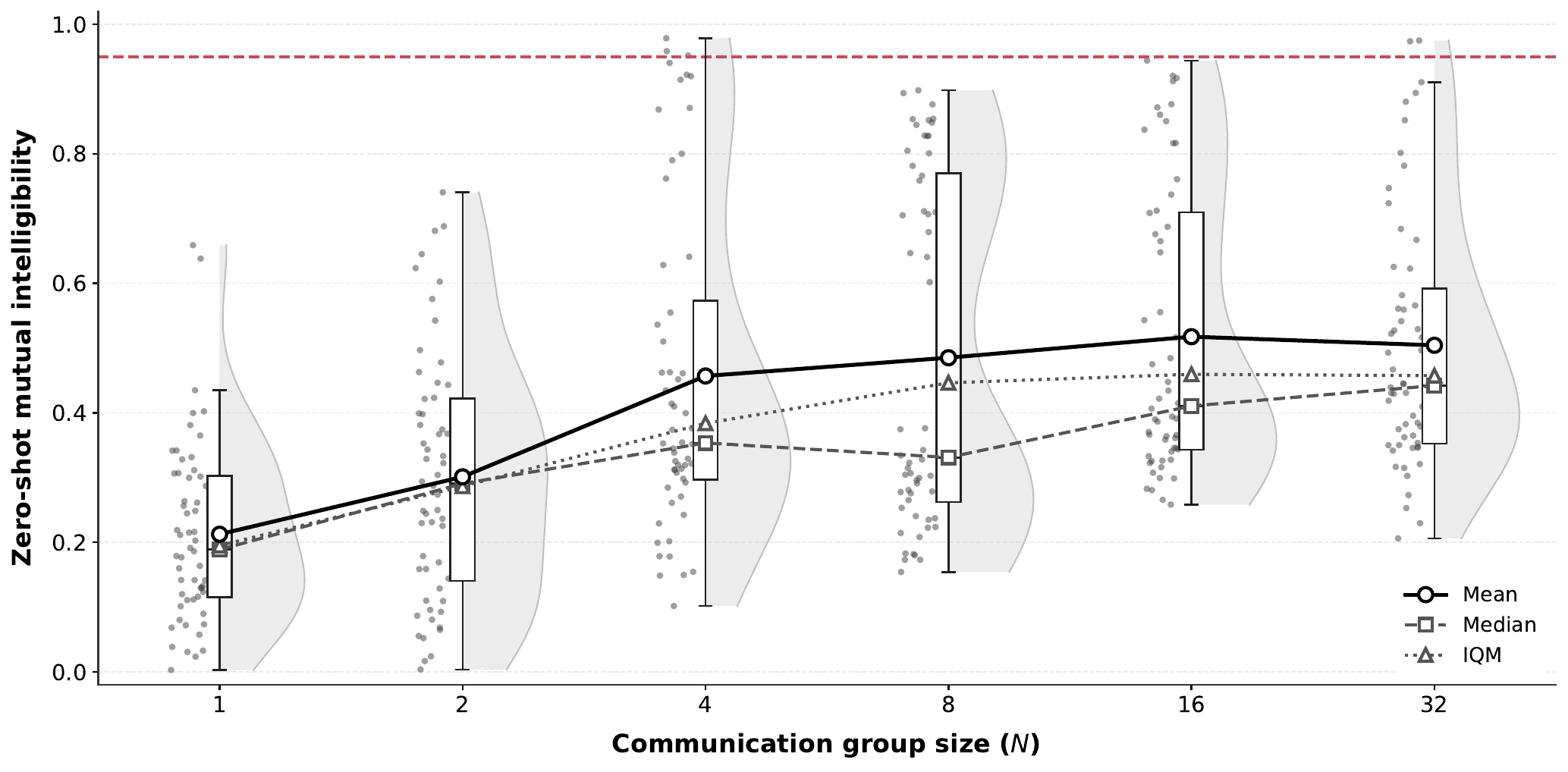}
        \caption{MNIST dataset: the number of strokes set to 10}
    \end{subfigure}
    \hfill
    \begin{subfigure}[b]{\textwidth}
        \centering
        \includegraphics[width=0.8\textwidth]{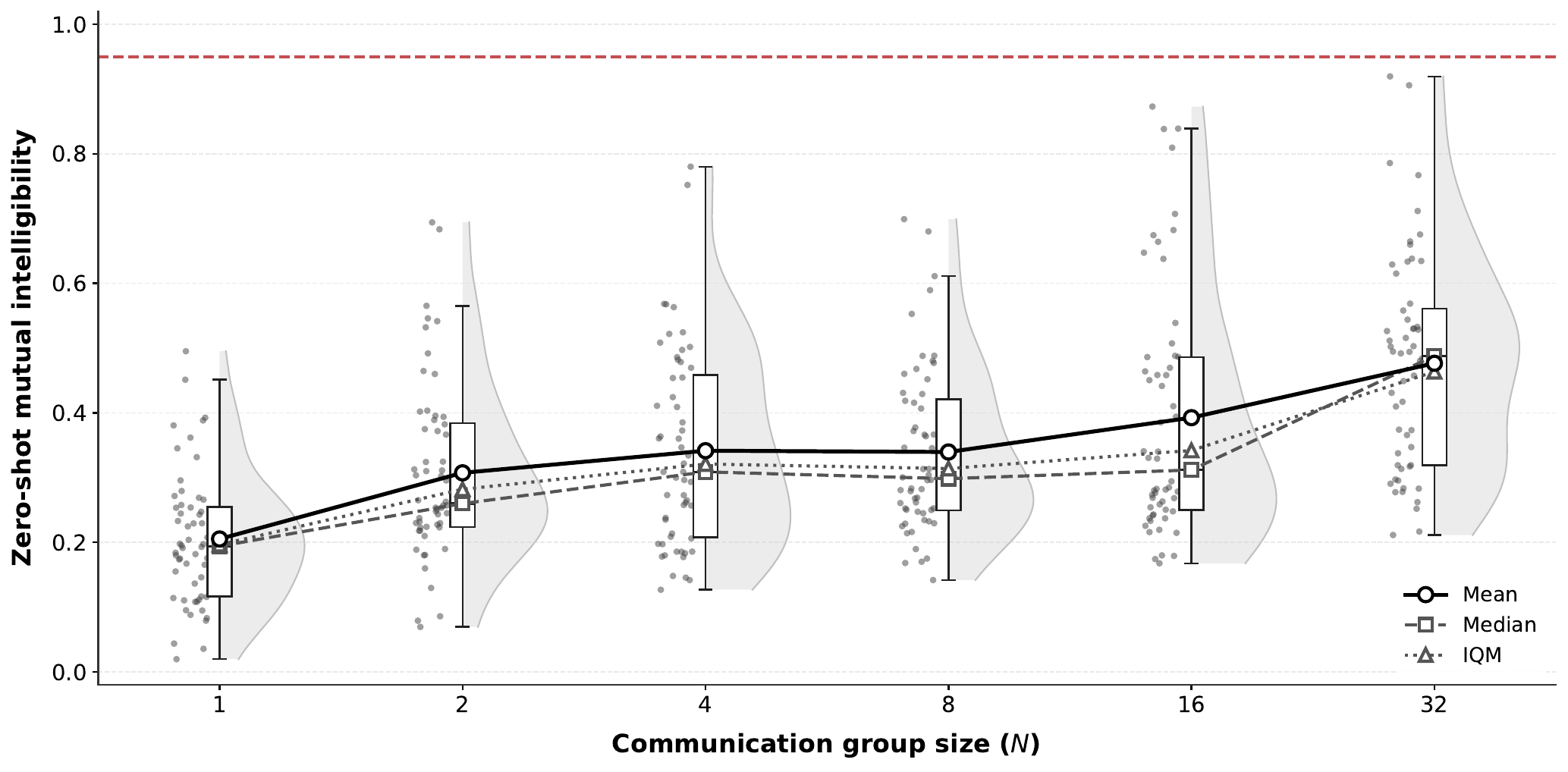}
        \caption{MNIST dataset: the number of strokes set to 30}
    \end{subfigure}
\caption{
    ZMI evaluated on the MNIST dataset when varying the number of strokes used to draw sketches used in drawing sketches.
    In the main text, the number of strokes is set to 20.
    In the top panel (a), the number of strokes is decreased to 10, whereas in the bottom panel (b), it is increased to 30.
    We observe that, for the MNIST dataset, the increase in ZMI with communication group size remains robust under both fewer and more strokes
    }
\label{fig:zmi_strokes_mnist}
\end{figure}

\begin{figure}[ht]
    \centering
    \begin{subfigure}[b]{\textwidth}
        \centering
        \includegraphics[width=0.8\textwidth]{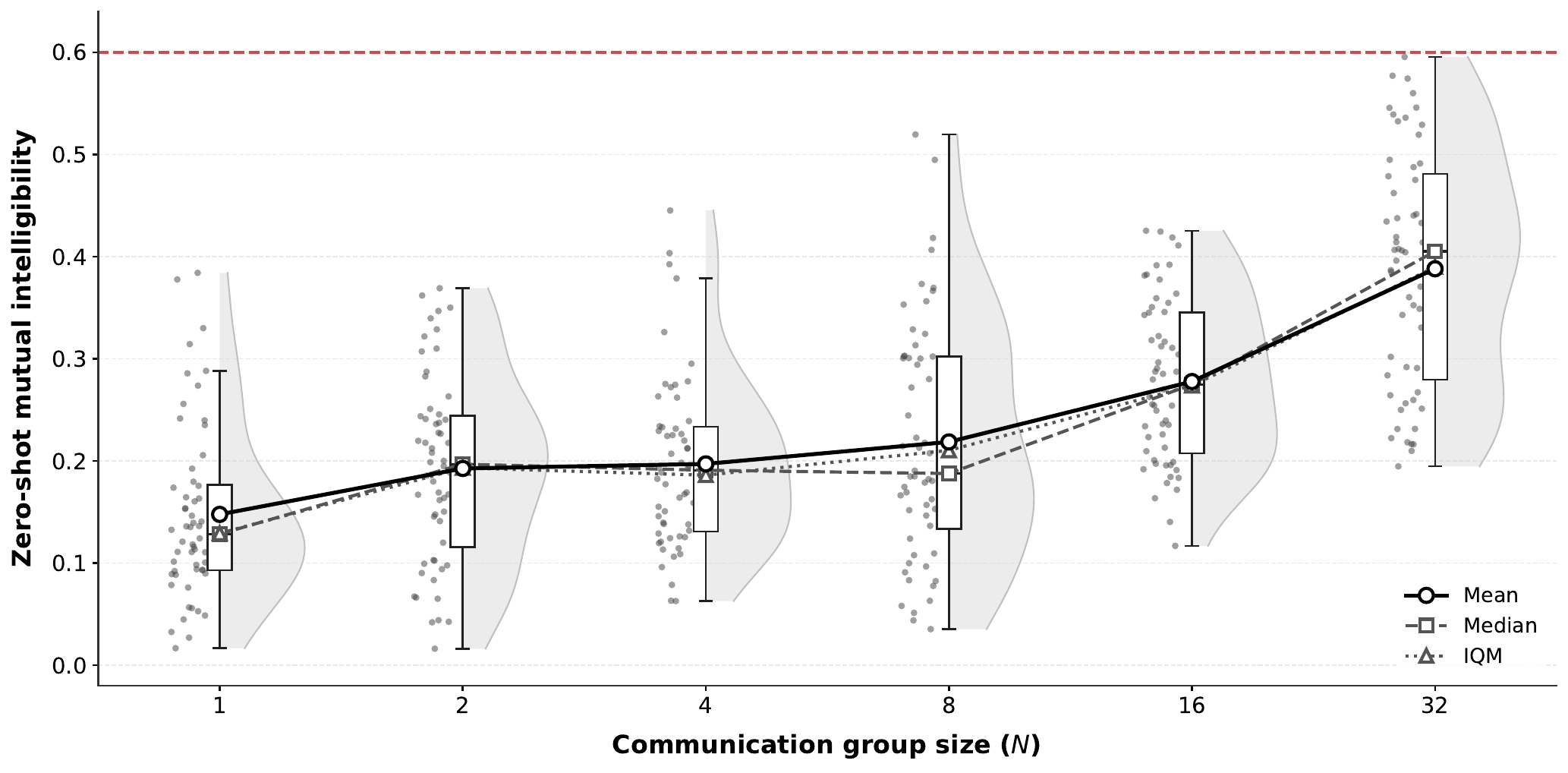}
        \caption{CIFAR-10 dataset: the number of strokes set to 10}
    \end{subfigure}
    \hfill
    \begin{subfigure}[b]{\textwidth}
        \centering
        \includegraphics[width=0.8\textwidth]{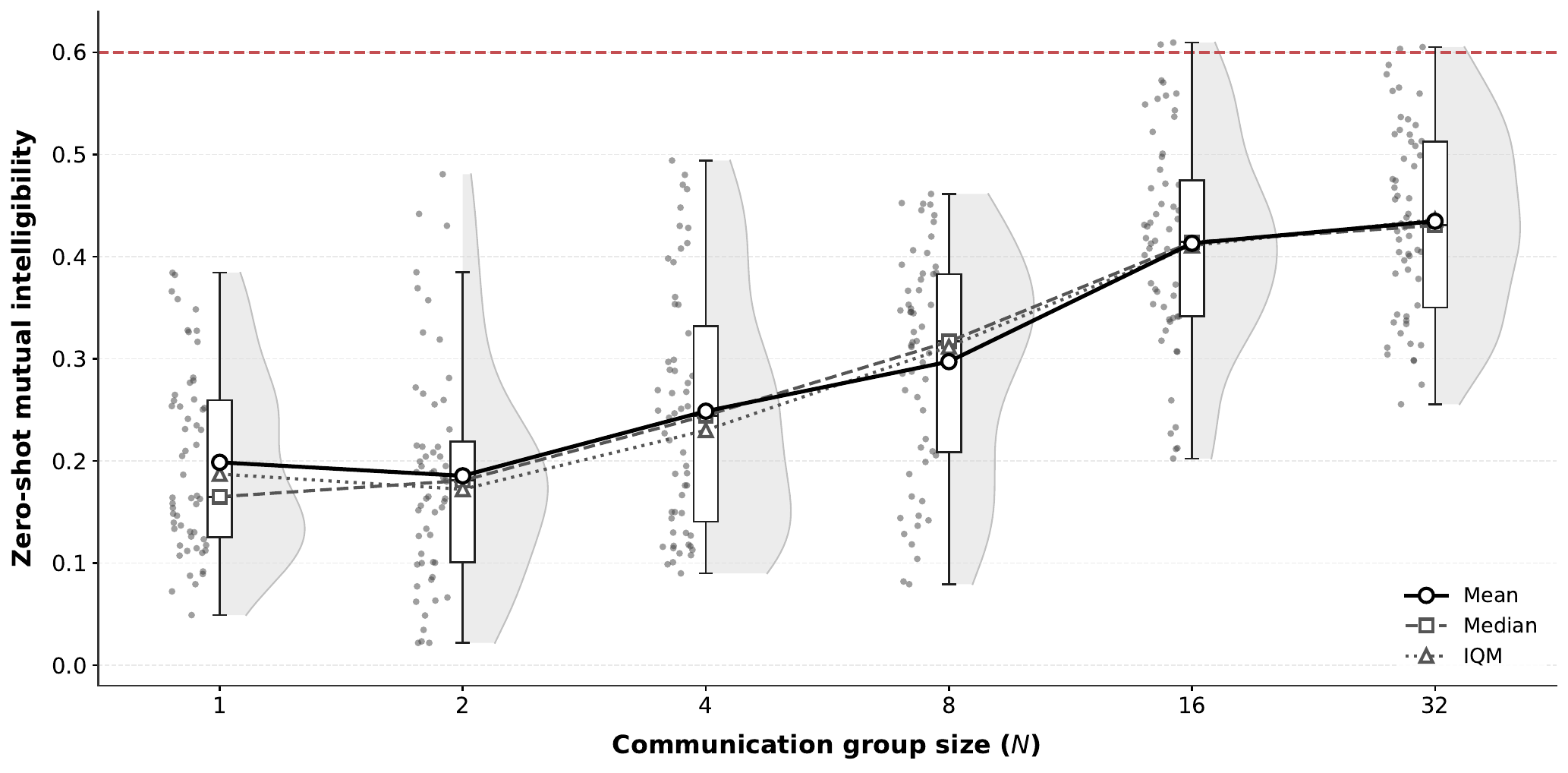}
        \caption{CIFAR-10 dataset: the number of strokes set to 30}
    \end{subfigure}
\caption{
    ZMI evaluated on the CIFAR-10 dataset when varying the number of strokes used to draw sketches used in drawing sketches.
    In the main text, the number of strokes is set to 20.
    In the top panel (a), the number of strokes is decreased to 10, whereas in the bottom panel (b), it is increased to 30.
    We observe that, for the CIFAR-10 dataset, the increase in ZMI with communication group size remains robust under both fewer and more strokes.
    }
\label{fig:zmi_strokes_cifar}
\end{figure}

The number of strokes controls the expressive capacity of the sketches produced by the sender.
A smaller stroke budget forces the sender to communicate using a more compressed visual signal, whereas a larger stroke budget allows the sender to produce more detailed sketches.
In the main experiments, the number of strokes is fixed to $20$.
To examine whether the observed population-scaling effect depends on this particular sketch capacity, we repeat the analysis with both fewer and more strokes.
Specifically, we evaluate stroke budgets of $10$ and $30$ for both MNIST and CIFAR-10.
\autoref{fig:zmi_strokes_mnist} and \autoref{fig:zmi_strokes_cifar} show the results for MNIST and CIFAR-10, respectively.
Across both stroke budgets, ZMI continues to increase with communication group size.
This suggests that the positive relationship between population size and ZMI is not specific to the default choice of $20$ strokes.
Even when the sketches are constrained to use fewer strokes, larger populations still lead to higher ZMI, and the same qualitative trend is preserved when the sketches are allowed to use more strokes.
\clearpage
\subsection{Decreasing the number of candidate images}
\begin{figure}[ht]
\centering
\begin{subfigure}[b]{\textwidth}
\centering
\includegraphics[width=0.8\textwidth]{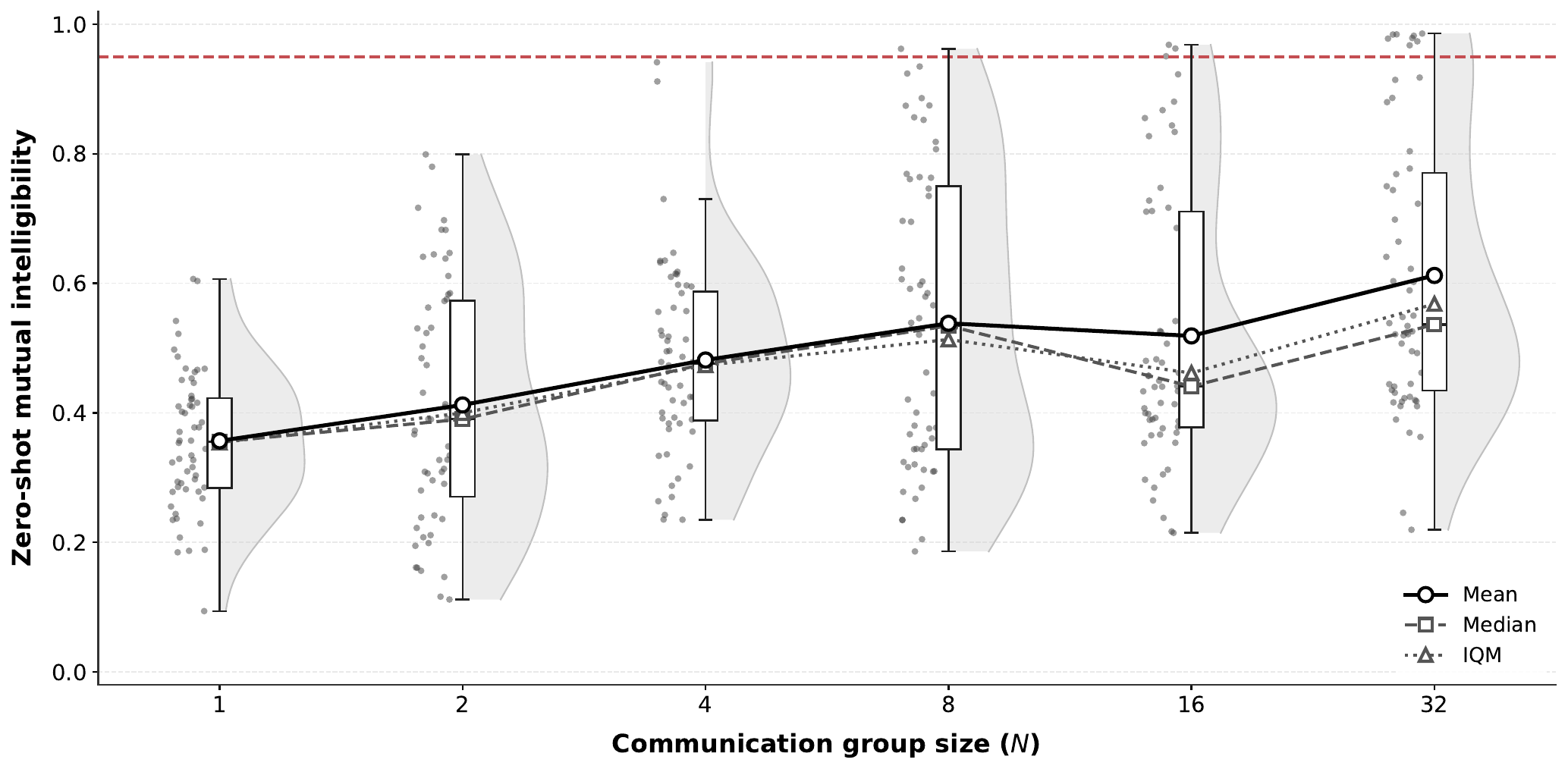}
\caption{MNIST dataset: the number of candidate images is set to 5}
\end{subfigure}

\begin{subfigure}[b]{\textwidth}
    \centering
    \includegraphics[width=0.8\textwidth]{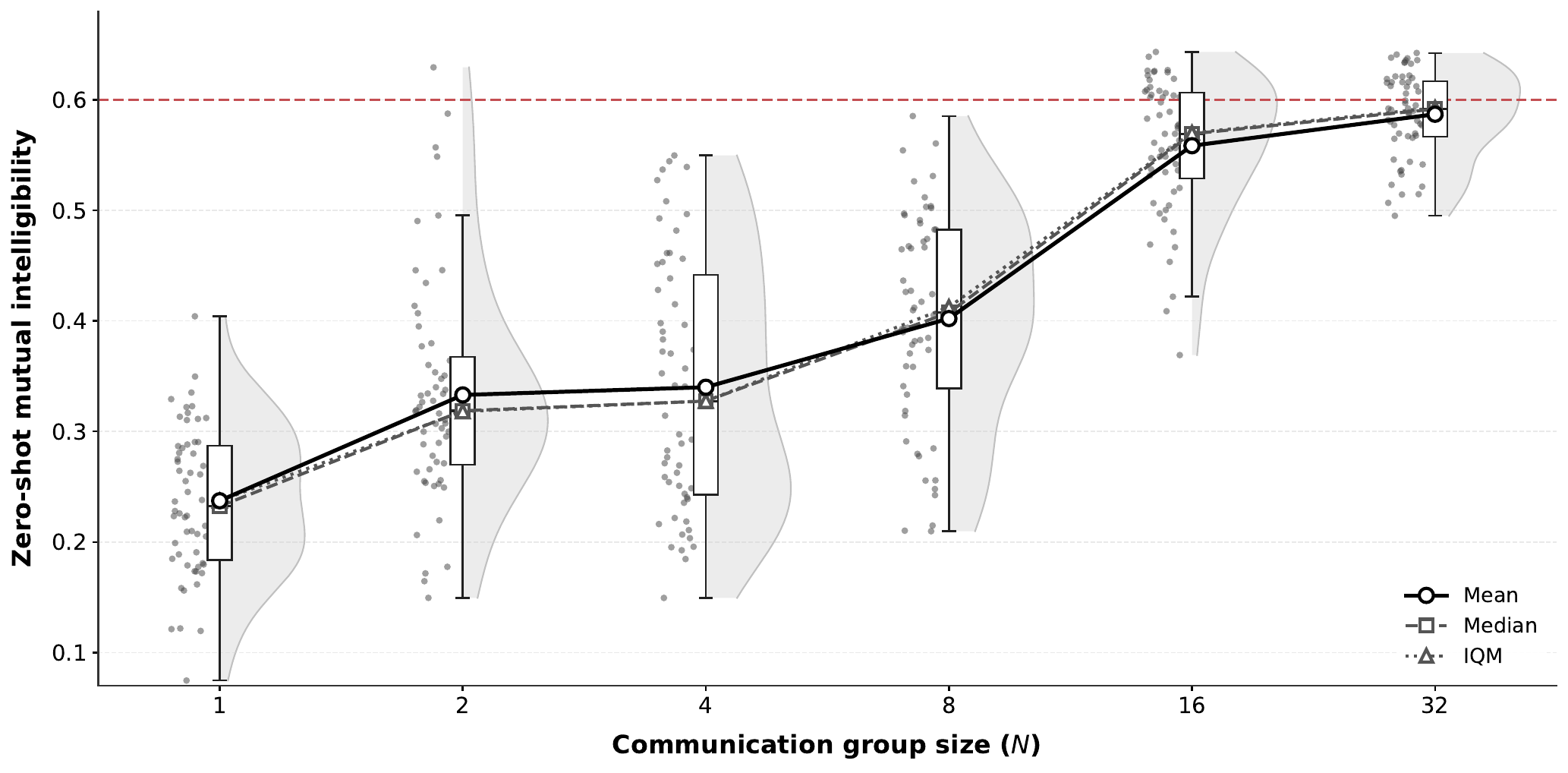}
    \caption{CIFAR-10 dataset: the number of candidate images is set to 5}
\end{subfigure}

\caption{
ZMI evaluated on the MNIST and CIFAR-10 datasets after decreasing the number of candidate images available to the receiver.
In the main text, the number of candidate images is set to 10.
Here, we decrease the number of candidate images to 5.
For both MNIST and CIFAR-10, the increase in ZMI with communication group size remains robust even when the receiver chooses from a smaller candidate set.
}
\label{fig:zmi_candidate_images}
\end{figure}

In the main experiments, the receiver selects the target image from $10$ candidate images, corresponding to the number of classes in MNIST and CIFAR-10.
This setting uses the largest candidate set available under our experimental design.
To examine whether the observed population-scaling effect depends on this choice, we repeat the analysis after decreasing the number of candidate images to $5$.
Reducing the candidate set changes the difficulty of the receiver's discrimination problem, since the receiver needs to identify the target among fewer alternatives.
\autoref{fig:zmi_candidate_images} shows the results for both MNIST and CIFAR-10.
For both datasets, ZMI continues to increase as the communication group size grows, even when the number of candidate images is reduced to $5$.
This suggests that the positive relationship between population size and ZMI is not specific to the default $10$-candidate setting used in the main experiments.
Together with the previous robustness analyses, these results indicate that the population-scaling effect remains stable across changes in the reference in-group validation accuracy, the sketch stroke budget, and the number of candidate images available to the receiver.
\clearpage
\section{Model-Scaling Ablation}
\label{app:model_scaling_ablation}
A possible alternative explanation for the population-scaling results is that the increase in zero-shot mutual intelligibility (ZMI) may be driven by model capacity rather than by population size.
This is a plausible concern, as prior work has shown that increasing model scale can improve generalization or transfer performance in large neural networks~\citep{kaplan2020scaling,zhai2022scaling}.
We therefore ask whether ZMI also improves when the sender and receiver networks are enlarged, but the communication population is not scaled.

To isolate the effect of model capacity, we fix the training population to a single sender and a single receiver, i.e., self-communication.
We then compare four settings: the default sender and receiver, an enlarged sender with the default receiver, the default sender with an enlarged receiver, and an enlarged sender paired with an enlarged receiver.
The default sender and receiver architectures are summarized in \autoref{tab:model_architecture}.
The enlarged models preserve the same input and output interfaces as the default models.
Specifically, the enlarged sender replaces the CandidateReceptor inside the sender with a wider CandidateReceptor, while keeping the final projection to $80$ sketch parameters unchanged.
The enlarged receiver replaces both the sketch encoder and the candidate-image encoder with larger counterparts, while keeping the final sketch--candidate matching MLP interface unchanged.
Thus, the sender still outputs $80$ parameters corresponding to $20$ strokes, and the receiver still maps a rasterized sketch and a candidate set to one logit per candidate.
The resulting model sizes are summarized in \autoref{tab:model_scaling_networks}.
\begin{table}[ht]
    \centering
    \small
    \renewcommand{\arraystretch}{2.0}
    \setlength{\tabcolsep}{6pt}
    \rowcolors{2}{gray!5}{white}
    \begin{tabular}{lll}
        \toprule
        Module & Operation & Output shape \\
        \midrule

        \multicolumn{3}{l}{\textit{Sender}} \\
        \midrule
        Input
        & Image
        & $(B,\mathrm{ch},h,w)$ \\

        Image encoder
        & CandidateReceptor (CNN)
        & $(B,256)$ \\

        Sketch head
        & Linear$(256,80)$, Sigmoid
        & $(B,80)$ \\

        Stroke parameters
        & Reshape $80 \rightarrow 20 \times 2 \times 2$
        & $(B,20,2,2)$ \\

        Sketch image
        & Rasterization
        & $(B,1,64,64)$ \\

        \midrule
        \multicolumn{3}{l}{\textit{Receiver}} \\
        \midrule
        Sketch input
        & Rasterized sketch
        & $(B,1,64,64)$ \\

        Sketch encoder
        & SketchReceptor (CNN)
        & $(B,256)$ \\

        Candidate input
        & Candidate images
        & $(B,C,\mathrm{ch},h,w)$ \\

        Candidate encoder
        & CandidateReceptor, applied per candidate
        & $(B,C,256)$ \\

        Joint embedding
        & Concatenate sketch and candidate embeddings
        & $(B,C,512)$ \\

        Matching head
        & Linear$(512,512)$, ReLU, Linear$(512,1)$
        & $(B,C)$ \\

        \bottomrule
    \end{tabular}
    \caption{
    Default sender and receiver architectures.
    The sender encodes an input image and predicts $80$ sketch parameters, corresponding to $20$ strokes with two endpoints and two coordinates per endpoint.
    The predicted strokes are rasterized into a $64 \times 64$ sketch image.
    The receiver encodes the rasterized sketch and each candidate image separately, concatenates the corresponding embeddings, and scores each sketch--candidate pair with an MLP.
    }
    \label{tab:model_architecture}
\end{table}
\begin{table}[t]
    \centering
    \renewcommand{\arraystretch}{1.35}
    \setlength{\tabcolsep}{7pt}
    \begin{tabular}{llrrr}
        \toprule
        Dataset & Agent & Default & Enlarged & Increase \\
        \midrule
        MNIST
        & Sender
        & $0.889$M
        & $4.267$M
        & $4.80\times$ \\

        MNIST
        & Receiver
        & $1.520$M
        & $6.835$M
        & $4.50\times$ \\

        CIFAR-10
        & Sender
        & $1.135$M
        & $5.252$M
        & $4.63\times$ \\

        CIFAR-10
        & Receiver
        & $1.766$M
        & $7.821$M
        & $4.43\times$ \\
        \bottomrule
    \end{tabular}
    \caption{
    Model sizes used in the model-scaling ablation.
    The enlarged models are substantially larger than the default models while preserving the same input and output interfaces.
    }
    \label{tab:model_scaling_networks}
\end{table}

\begin{figure}[ht]
\centering
\begin{subfigure}[b]{\textwidth}
\centering
\includegraphics[width=0.8\textwidth]{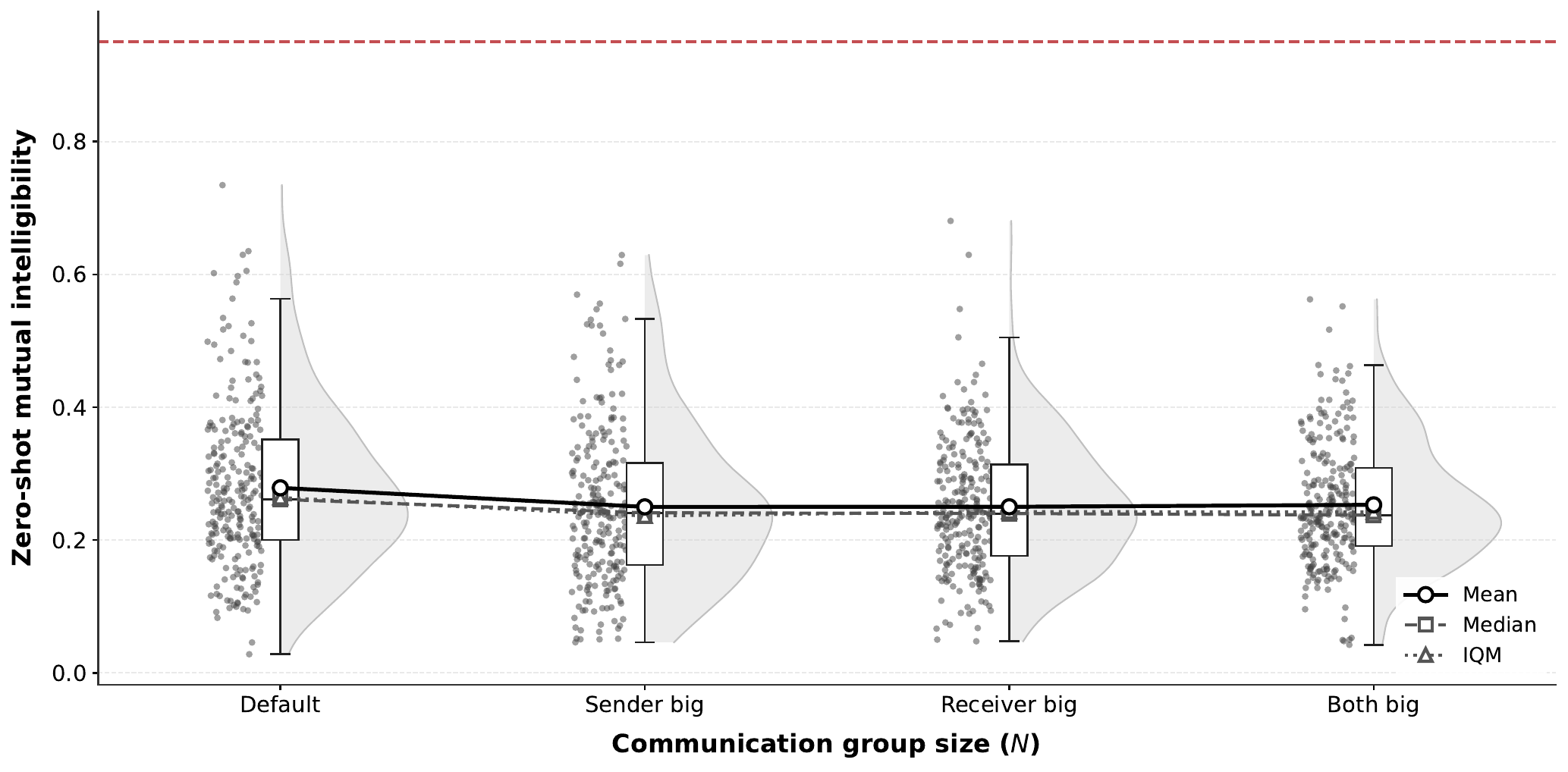}
\caption{MNIST dataset: enlarging the sender network, receiver network, or both}
\end{subfigure}
\begin{subfigure}[b]{\textwidth}
    \centering
    \includegraphics[width=0.8\textwidth]{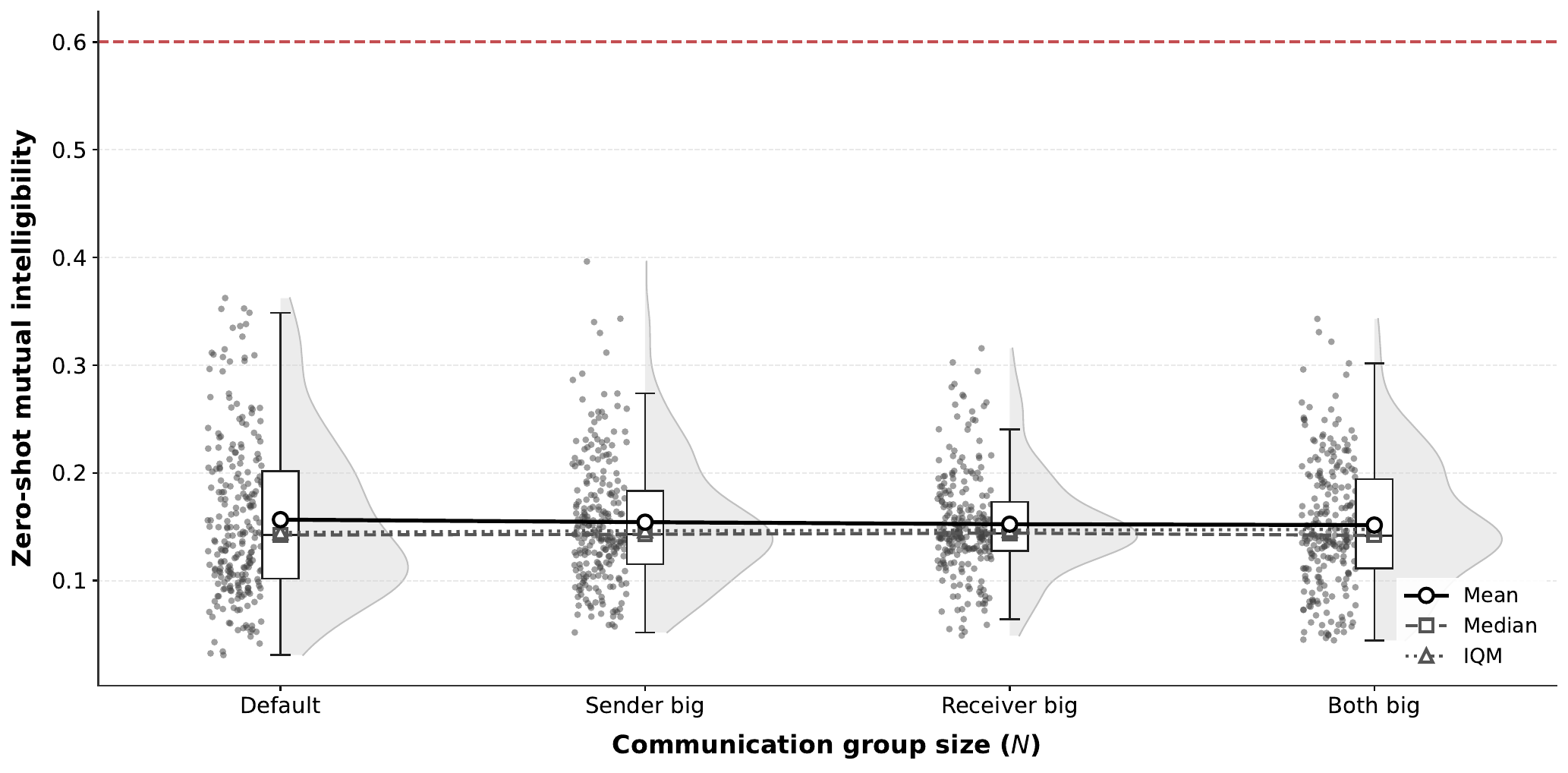}
    \caption{CIFAR-10 dataset: enlarging the sender network, receiver network, or both}
\end{subfigure}

\caption{
    Model-scaling ablation on MNIST and CIFAR-10.
    In all settings, the communication population is fixed to one sender and one receiver.
    We compare the default architecture against three enlarged variants: enlarged sender only, enlarged receiver only, and enlarged sender and receiver together.
    For both datasets, increasing model capacity without increasing the number of communication partners does not produce a systematic increase in zero-shot mutual intelligibility (ZMI).
}
\label{fig:zmi_big}
\end{figure}

\autoref{fig:zmi_big} shows the results for the four capacity settings summarized in \autoref{tab:model_scaling_networks}.
For both MNIST and CIFAR-10, enlarging the sender, the receiver, or both does not produce a systematic increase in ZMI.
Instead, ZMI remains approximately unchanged across all four model-capacity settings.
This contrasts with the main population-scaling experiments, where increasing the number of interacting senders and receivers consistently improves ZMI.
Thus, the observed improvement in zero-shot mutual intelligibility is not explained by model scaling alone.
Rather, the results support the interpretation that population scaling itself promotes communication protocols that are more mutually intelligible across independently trained agents.

\clearpage
\section{Additional results on one-to-many communications}
\label{app:linearity}
\begin{figure}[ht]
    \centering
    \begin{subfigure}[b]{0.4\textwidth}
        \centering
        \includegraphics[width=\textwidth]{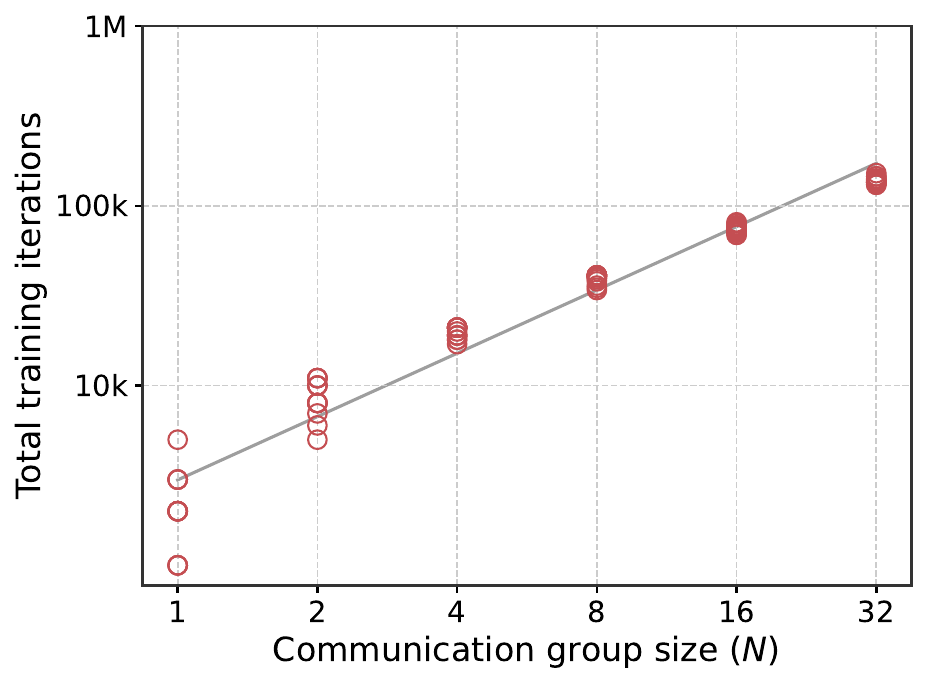}        
        \caption{One to many communications: MNIST dataset}
    \end{subfigure}
    \hspace{0.05\textwidth}
    \begin{subfigure}[b]{0.4\textwidth}
        \centering
        \includegraphics[width=\textwidth]{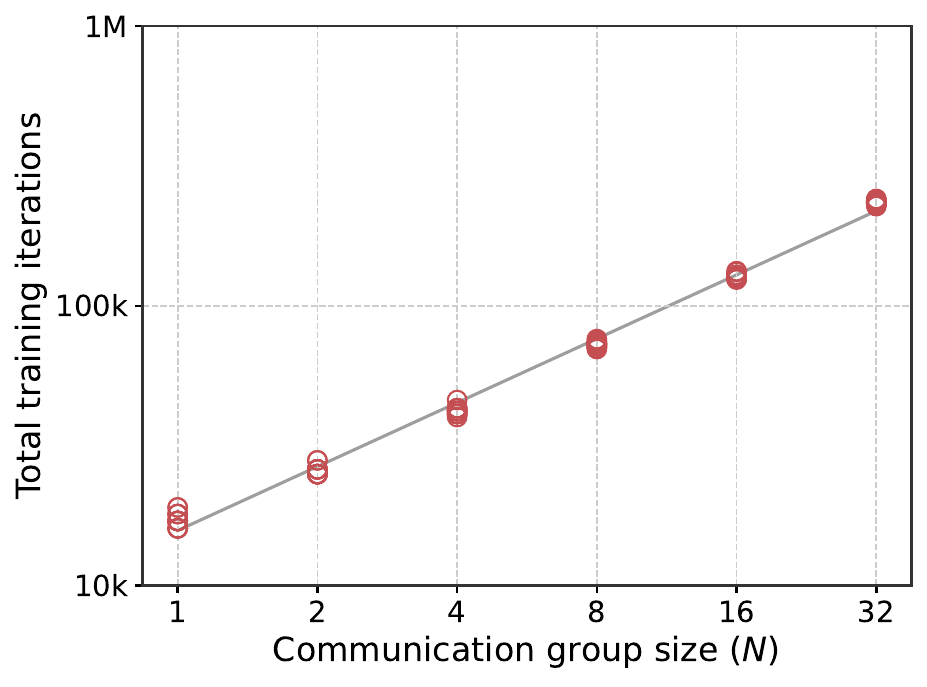}        
        \caption{One to many communications: CIFAR-10 dataset}
    \end{subfigure}
\caption{
    Population scaling increases training cost approximately linearly, even for one-to-many communications, for both the MNIST and the CIFAR-10 datasets.
    }
\label{fig:onetomany_linearity}
\end{figure}
\clearpage
\section{Details on differentiable rasterization method}
\label{app:rasterization}
The sender outputs a set of line segments parameterized by their endpoints.
In our experiments, each sender output consists of $20$ strokes, represented as an $80$-dimensional vector and reshaped into $20 \times 2 \times 2$, where each stroke has two endpoints and each endpoint has two normalized coordinates.
These continuous stroke parameters are converted into a rasterized sketch using a differentiable rasterization procedure originally proposed by \citet{mihai2021differentiable} and used in the sketch-based emergent-communication setting of \citet{mihai2021learning}.

For each output pixel, we compute its normalized pixel-center coordinate and measure its squared distance to each line segment.
This is done by projecting the pixel center onto the corresponding segment, clipping the projection coefficient to the segment interval, and computing the distance to the closest point on the segment.
The distance is then converted into a soft stroke intensity using a Gaussian kernel.
Multiple strokes are composited multiplicatively, implemented in log space for numerical stability.
Because all operations are differentiable with respect to the stroke endpoints, gradients can propagate from the receiver loss through the rasterized sketch back to the sender.
The implementation additionally caches pixel grids, processes large batches in chunks, and rescales the kernel width with the image resolution so that the effective stroke thickness remains approximately constant across rasterization sizes.
Please refer to the attached code for the actual implementation.
\clearpage
\section{Training and implementation details}
\label{app:training_details}
This section summarizes the main training, task, and implementation configurations used in our experiments.
The task-level setup is given in \autoref{tab:task_config}, and the optimization hyperparameters are summarized in \autoref{tab:training_hyperparams}.
In all experiments, the sender predicts continuous stroke parameters, which are rasterized into a sketch and passed to the receiver together with candidate images.
During training, a sender--receiver pair is sampled at each iteration from the communication population, and both agents in the sampled pair are updated using the cross-entropy loss over candidate images.

\begin{table}[ht]
    \centering
    \renewcommand{\arraystretch}{1.5}
    \setlength{\tabcolsep}{7pt}
    \rowcolors{2}{gray!5}{white}
    \begin{tabular}{ll}
        \toprule
        Item & Setting \\
        \midrule
        Datasets & MNIST, CIFAR-10 \\
        Image resolution & $28 \times 28$ for MNIST; $32 \times 32$ for CIFAR-10 \\
        Image channels & $1$ for MNIST; $3$ for CIFAR-10 \\
        Number of classes & $10$ \\
        Candidate images & $10$ candidates per example \\
        Candidate construction & One image sampled from each class, randomly permuted \\
        Sender output & $80$ parameters $=20 \times 2 \times 2$ \\
        Number of strokes & $20$ \\
        Rasterized sketch size & $64 \times 64$ \\
        Receiver output & One logit per candidate \\
        Loss & Cross-entropy over candidate indices \\
        MNIST normalization & Mean $0.1308$, std. $0.3081$ \\
        CIFAR-10 normalization & Mean $(0.4914,0.4822,0.4465)$; std. $(0.2023,0.1994,0.2010)$ \\
        CIFAR-10 augmentation & Random crop with padding $4$; horizontal flip with probability $0.5$ \\
        \bottomrule
    \end{tabular}
    \caption{
    Task and data configuration.
    }
    \label{tab:task_config}
\end{table}

\begin{table}[ht]
    \centering
    \renewcommand{\arraystretch}{1.5}
    \setlength{\tabcolsep}{7pt}
    \rowcolors{2}{gray!5}{white}
    \begin{tabular}{ll}
        \toprule
        Hyperparameter & Setting \\
        \midrule
        Batch size & $64$ \\
        Evaluation interval & Every $1{,}000$ pair updates \\
        Evaluation batches & $100$ \\
        Optimizer & AdamW \\
        Sender learning rate & $10^{-4}$ \\
        Receiver learning rate & $10^{-4}$ \\
        Early-stopping threshold & $0.95$ for MNIST; $0.60$ for CIFAR-10 \\
        Early-stopping metric & Moving average of recent validation communication accuracy \\
        \bottomrule
    \end{tabular}
    \caption{
    Optimization hyperparameters.
    The training loop runs for up to $N$ times the maximum number of updates, where $N$ is the communication group size, so that each sender and receiver receives approximately the specified number of updates in expectation.
    Training is stopped early once the recent validation communication accuracy exceeds the dataset-specific threshold.
    }
    \label{tab:training_hyperparams}
\end{table}
\clearpage
\section{Computational Infrastructure}
\label{app:compute_infrastructure}
The main experiments were run on a server equipped with eight NVIDIA RTX PRO 6000 Blackwell GPUs.
Each experimental group corresponds to an independently trained communication population, initialized with a different random seed.
In the main experiments, we use $M=16$ independently trained groups.
The groups are distributed across the eight GPUs, with each GPU assigned two groups.
This parallelization is used only to reduce wall-clock time: groups are trained independently, with no parameter sharing, gradient exchange, or communication across GPUs.
\autoref{tab:compute_infrastructure} summarizes the details.
\begin{table}[ht]
    \centering
    \renewcommand{\arraystretch}{1.3}
    \setlength{\tabcolsep}{7pt}
    \begin{tabular}{ll}
        \toprule
        Item & Setting \\
        \midrule
        GPU hardware & $8 \times$ NVIDIA RTX PRO 6000 Blackwell \\
        PyTorch version & 2.12.0+cu130 \\
        CUDA version & 13.1 \\
        Experimental groups & $M=16$ independently trained groups \\
        Cross-GPU communication & None \\
        \bottomrule
    \end{tabular}
    \caption{
    Computational infrastructure used for the main experiments.
    }
    \label{tab:compute_infrastructure}
\end{table}
\clearpage
\section{Creating Perceptually Grounded Sketches}
\label{app:perceptual_sketches}
The procedure of creating perceptually grounded sketches is motivated by the work of \citet{mihai2021learning}.
However, in their work, the perceptual loss is added on top of the emergent communication loss, so that sketches are optimized both for communication and for perceptual similarity to the input image.
Here, we use the perceptual objective by itself, without the receiver or communication loss, to visualize what the sender architecture can produce when directly encouraged to preserve perceptual information from the input image.

\begin{figure}[ht]
\centering
\begin{subfigure}[b]{\textwidth}
\centering
\includegraphics[width=0.8\textwidth]{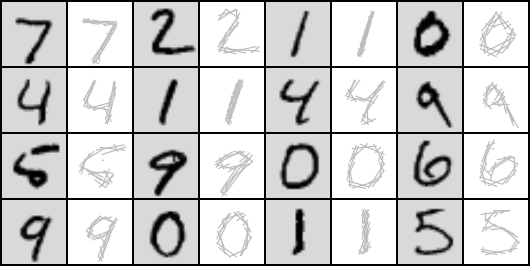}
\caption{MNIST dataset}
\end{subfigure}
\begin{subfigure}[b]{\textwidth}
    \centering
    \includegraphics[width=0.8\textwidth]{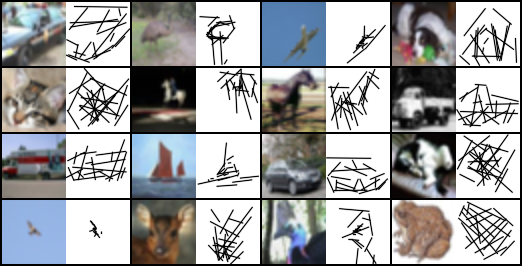}
    \caption{CIFAR-10 dataset}
\end{subfigure}

\caption{
    Examples of perceptually grounded sketches.
}
\label{fig:perceptual_examples}
\end{figure}

Given an input image $P$, the sender predicts $20$ strokes, which are rasterized into a sketch image $S$ using the differentiable rasterizer described in \autoref{app:rasterization}.
The perceptual loss compares $P$ and $S$ in the feature space of a frozen image classifier.
For CIFAR-10, we use a pretrained CIFAR-10 VGG16-BN model, whereas for MNIST, we train a VGG-style classifier on MNIST and use its intermediate feature blocks.
At each selected layer, we normalize the feature activations channel-wise and compute the squared distance between the normalized features of the original image and the sketch.
The final perceptual loss is the sum of these feature-level distances across layers.
The sender is then trained to minimize this perceptual loss.
\autoref{fig:perceptual_examples} shows examples of sketches produced by senders trained with this perceptual objective.
Note that for the MNIST dataset,  the background is black and the strokes are white.
Note that, to generate perceptually grounded sketches, we inverted the colors of the background and the strokes to ensure consistency with the colors used in the rasterization method.
\begin{table}[ht]
    \centering
    \renewcommand{\arraystretch}{1.5}
    \setlength{\tabcolsep}{7pt}
    \rowcolors{2}{gray!5}{white}
    \begin{tabular}{lll}
        \toprule
        Item & MNIST & CIFAR-10 \\
        \midrule
        Sender output & $20$ strokes & $20$ strokes \\
        Sketch resolution & $64 \times 64$ & $64 \times 64$ \\
        Loss image size & $28 \times 28$ & $32 \times 32$ \\
        Perceptual backbone & VGG-style MNIST classifier & CIFAR-10 VGG16-BN \\
        Feature blocks & $[{:}4]$, $[4{:}9]$, $[9{:}14]$ & $[{:}23]$, $[23{:}33]$, $[33{:}43]$ \\
        Backbone parameters & Frozen & Frozen \\
        Optimizer & AdamW & AdamW \\
        Learning rate & $10^{-4}$ & $10^{-4}$ \\
        Batch size & $128$ & $128$ \\
        Maximum iterations & $100{,}000$ & $1{,}000{,}000$ \\
        Checkpoint selection & Best validation perceptual loss & Best validation perceptual loss \\
        \bottomrule
    \end{tabular}
    \caption{
    Configuration for training perceptually grounded sketch generators.
    The sender is trained without a receiver, using only the perceptual loss between the input image and the rasterized sketch.
    The perceptual backbone is kept frozen throughout sender training.
    }
    \label{tab:perceptual_sketch_config}
\end{table}
\clearpage
\section{Examples of the created sketches}
\label{app:examples}
\begin{figure}[ht]
    \centering
    \includegraphics[width=0.6\textwidth]{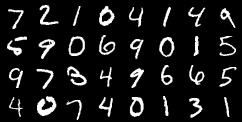}        
    \caption{
        Example MNIST images.
        }
    \label{fig:mnist_example}
\end{figure}
\begin{figure}[ht]
    \centering
    \includegraphics[width=0.6\textwidth]{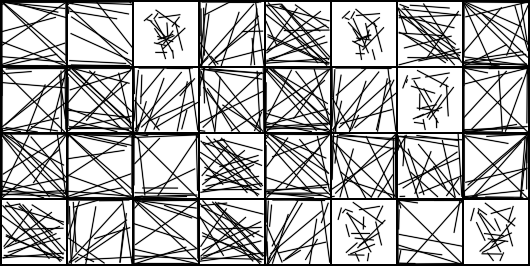}        
    \caption{
        Create sketches from the MNIST data (\autoref{fig:mnist_example}) by agent in group size $N = 1$.
        }
\end{figure}
\begin{figure}[ht]
    \centering
    \includegraphics[width=0.6\textwidth]{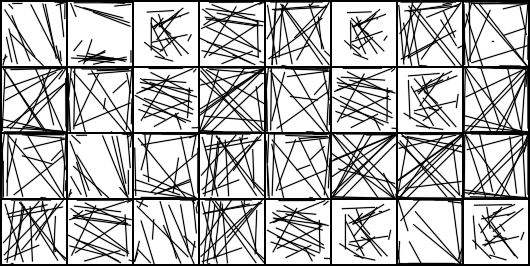}        
    \caption{
        Create sketches from the MNIST data (\autoref{fig:mnist_example}) by agent in group size $N = 2$.
        }
\end{figure}
\begin{figure}[ht]
    \centering
    \includegraphics[width=0.6\textwidth]{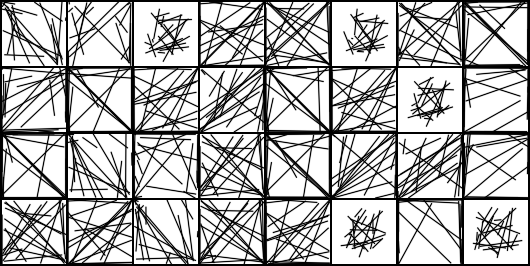}        
    \caption{
        Create sketches from the MNIST data (\autoref{fig:mnist_example}) by agent in group size $N = 4$.
        }
\end{figure}
\begin{figure}[ht]
    \centering
    \includegraphics[width=0.6\textwidth]{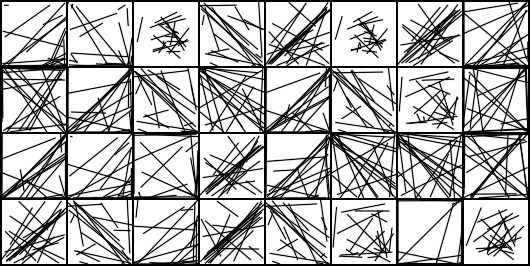}        
    \caption{
        Create sketches from the MNIST data (\autoref{fig:mnist_example}) by agent in group size $N = 8$.
        }
\end{figure}
\begin{figure}[ht]
    \centering
    \includegraphics[width=0.6\textwidth]{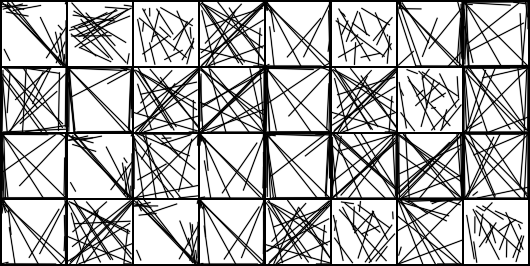}        
    \caption{
        Create sketches from the MNIST data (\autoref{fig:mnist_example}) by agent in group size $N = 16$.
        }
\end{figure}
\begin{figure}[ht]
    \centering
    \includegraphics[width=0.6\textwidth]{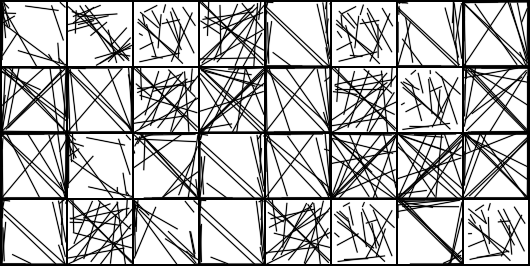}        
    \caption{
        Create sketches from the MNIST data (\autoref{fig:mnist_example}) by agent in group size $N = 32$.
        }
\end{figure}
\begin{figure}[ht]
    \centering
    \includegraphics[width=0.6\textwidth]{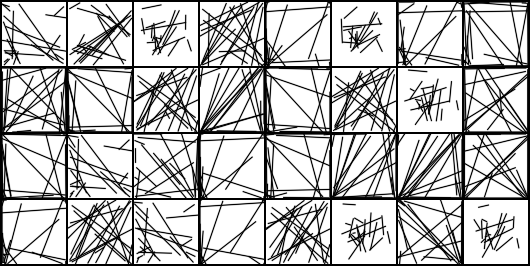}        
    \caption{
        Create sketches from the MNIST data (\autoref{fig:mnist_example}) by agent in group size $N = 64$.
        }
\end{figure}
\begin{figure}[ht]
    \centering
    \includegraphics[width=0.6\textwidth]{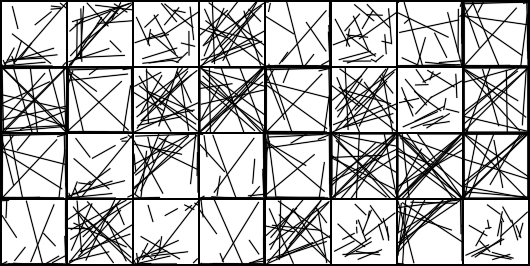}        
    \caption{
        Create sketches from the MNIST data (\autoref{fig:mnist_example}) by agent in group size $N = 128$.
        }
\end{figure}
\begin{figure}[ht]
    \centering
    \includegraphics[width=0.6\textwidth]{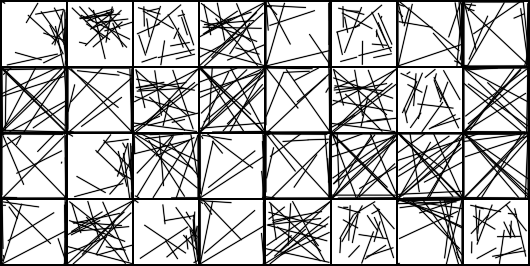}        
    \caption{
        Create sketches from the MNIST data (\autoref{fig:mnist_example}) by agent in group size $N = 256$.
        }
\end{figure}
\clearpage
\begin{figure}[ht]
    \centering
    \includegraphics[width=0.6\textwidth]{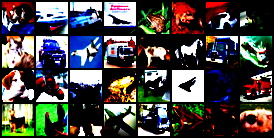}        
    \caption{
        Example CIFAR10 images.
        }
    \label{fig:CIFAR10_example}
\end{figure}
\begin{figure}[ht]
    \centering
    \includegraphics[width=0.6\textwidth]{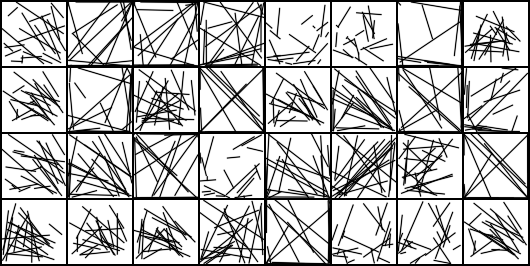}        
    \caption{
        Create sketches from the CIFAR10 data (\autoref{fig:CIFAR10_example}) by agent in group size $N = 1$.
        }
\end{figure}
\begin{figure}[ht]
    \centering
    \includegraphics[width=0.6\textwidth]{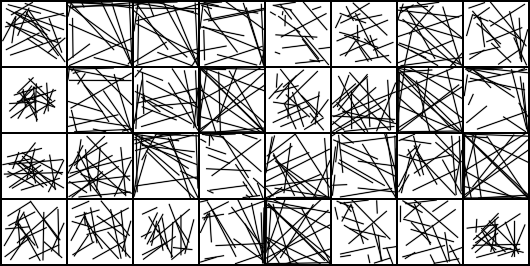}        
    \caption{
        Create sketches from the CIFAR10 data (\autoref{fig:CIFAR10_example}) by agent in group size $N = 2$.
        }
\end{figure}
\begin{figure}[ht]
    \centering
    \includegraphics[width=0.6\textwidth]{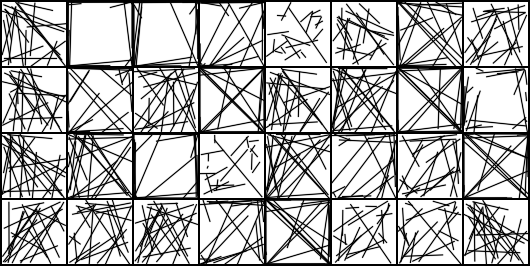}        
    \caption{
        Create sketches from the CIFAR10 data (\autoref{fig:CIFAR10_example}) by agent in group size $N = 4$.
        }
\end{figure}
\begin{figure}[ht]
    \centering
    \includegraphics[width=0.6\textwidth]{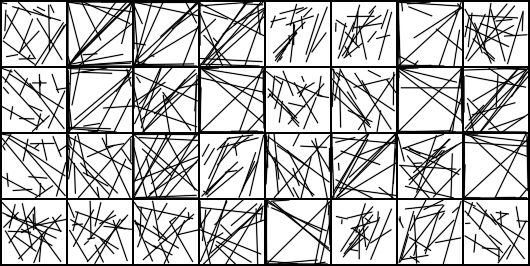}        
    \caption{
        Create sketches from the CIFAR10 data (\autoref{fig:CIFAR10_example}) by agent in group size $N = 8$.
        }
\end{figure}
\begin{figure}[ht]
    \centering
    \includegraphics[width=0.6\textwidth]{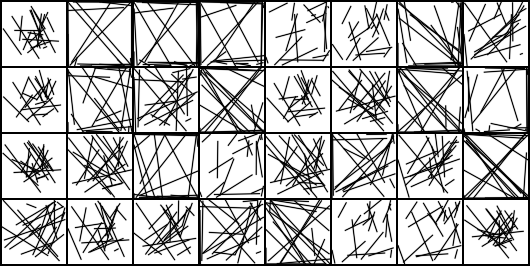}        
    \caption{
        Create sketches from the CIFAR10 data (\autoref{fig:CIFAR10_example}) by agent in group size $N = 16$.
        }
\end{figure}
\begin{figure}[ht]
    \centering
    \includegraphics[width=0.6\textwidth]{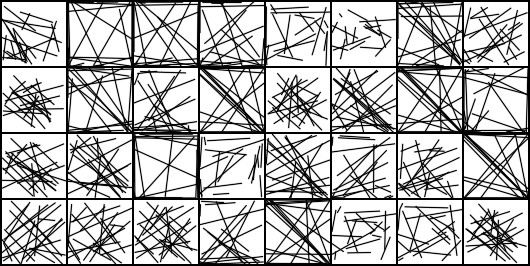}        
    \caption{
        Create sketches from the CIFAR10 data (\autoref{fig:CIFAR10_example}) by agent in group size $N = 32$.
        }
\end{figure}
\begin{figure}[ht]
    \centering
    \includegraphics[width=0.6\textwidth]{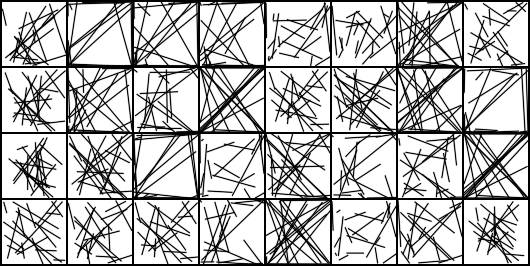}        
    \caption{
        Create sketches from the CIFAR10 data (\autoref{fig:CIFAR10_example}) by agent in group size $N = 64$.
        }
\end{figure}
\end{document}